%% file: main_arxiv.tex
\documentclass[twoside,11pt]{article}

\usepackage[preprint]{jmlr2e}

\usepackage{natbib}

\input{lib/preamble}

\input{lib/macros}
\input{lib/math_macros}

\usepackage{lastpage}

\jmlrheading{23}{2022}{1-\pageref{LastPage}}{1/21; Revised 5/22}{9/22}{21-0000}{Thijs Vogels and Hadrien Hendrikx and Martin Jaggi}

\ShortHeadings{Beyond spectral gap}{Vogels, Hendrikx, Jaggi}
\firstpageno{1}

\begin{document}
\title{Beyond spectral gap (extended): \\
  The role of the topology in decentralized learning}

\author{%
  \name Thijs Vogels* \email thijs.vogels@epfl.ch%
  \AND
  \name Hadrien Hendrikx* \email hadrien.hendrikx@epfl.ch%
  \AND
  \name Martin Jaggi \email martin.jaggi@epfl.ch%
  \AND
  \addr Machine Learning and Optimization Laboratory\\
  EPFL\\
  Lausanne, Switzerland
}

\editor{???}

\maketitle

\input{010_abstract.tex}


\input{020_introduction}
\input{030_related_work}
\input{040_random_quadratics}
\input{060_theory}
\input{065_experiments}
\input{070_conclusion}

\input{080_acknowledgements}

\vskip 0.2in

\bibliography{bibliography_autogen.bib}

\end{document}

%% file: lib/preamble.tex
\usepackage[utf8]{inputenc} 
\usepackage[T1]{fontenc}    
\usepackage{microtype}      
\usepackage{xspace}         
\usepackage{ifthen}

\usepackage{xargs}          

\usepackage{mdframed}

\usepackage{amssymb,amsmath,amsfonts}
\usepackage{bm}             
\usepackage{mathtools}
\usepackage{dsfont}

\usepackage{algorithm}
\usepackage{algpseudocode}  
\usepackage{booktabs}
\usepackage{tabularx}       

\usepackage[dvipsnames]{xcolor} 
\usepackage{graphicx}
\usepackage{tikz}           

\usetikzlibrary{patterns}
\usetikzlibrary{positioning}

\usepackage[toc, page, header]{appendix}
\usepackage{url}

\usepackage[english]{babel}
\addto\extrasenglish{
  
}

\usepackage{hyperref}       
\hypersetup{
    colorlinks=true,
    linkcolor=tab10_blue,
    citecolor=tab10_blue,
    filecolor=tab10_blue
}

%% file: lib/macros.tex
\newcommand{\eg}{e.g.\xspace}
\newcommand{\ie}{i.e.\xspace}
\newcommand{\iid}{i.i.d.\ }

\newcommand{\nocolor}[1]{}

\newcommand{\dims}{{{\nocolor{tab10_red}d}}}
\newcommand{\nwrk}{{{\nocolor{tab10_purple}n}}}
\newcommand{\lr}{{{\nocolor{tab10_red}\eta}}}
\newcommand{\gossip}{{{\nocolor{tab10_purple}\mW}}}
\newcommand{\weigval}{{{\nocolor{tab10_purple}\lambda}}}

\newcommand{\gossipweight}[1]{{{\nocolor{tab10_purple}w_{#1}}}}
\newcommand{\param}{{{\nocolor{tab10_orange}\xx}}}
\newcommand{\grad}{{{\nocolor{tab10_orange}\gg}}}

\newcommand{\noise}{{{\nocolor{tab10_green}\zeta}}}
\newcommand{\smoothf}{{{\nocolor{tab10_blue}L}}}
\newcommand{\randdata}{{{\nocolor{tab10_blue}\dd}}}

\newcommand{\mmat}{{{\nocolor{tab10_olive}\mM}}}
\newcommand{\mweight}[1]{{{\nocolor{tab10_olive}\mmat_{#1}}}}
\newcommand{\rate}{{{\nocolor{tab10_brown}r}}}
\newcommand{\LM}{{{\nocolor{tab10_pink}\mL_\mmat}}}
\newcommand{\LW}{{{\nocolor{tab10_pink}\mL_{\mW}}}}
\newcommand{\wcons}{{{\nocolor{tab10_gray}\omega}}}
\newcommand{\decay}{{{\nocolor{tab10_cyan}\gamma}}}

\newcommand{\neff}{{\nocolor{tab10_olive}{n_\gossip(\decay)}}}

\newcommand{\lmax}{{{\nocolor{tab10_orange}\weigval_2}}}

\newcommand{\esp}[1]{\expect\left[#1\right]}

\newcommand{\hetero}{\Delta^2_\gossip}
\newcommand{\spd}{{d_s}}
\newcommand{\xt}{\param\at{t}}
\newcommand{\xstar}{\param\at{\star}}
\newcommand{\xstarmat}{\param^\star_\lr}
\newcommand{\xstarmatbar}{\overline{\param^\star_\lr}}
\newcommand{\dxstramat}{\frac{\dd\xstarmat}{\dd \lr} }

\newcommand{\mzero}{\neff^{-1}}
\newcommand{\lmwsmooth}{\beta(\decay)}

\newcommand{\CG}{c_s}

\newcommand{\rw}{{{\nocolor{tab10_orange}\zz}}}
\newcommand{\rwo}{{{\nocolor{tab10_orange}\yy}}}
\newcommand{\rwn}{{{\nocolor{tab10_orange}\bm{\xi}}}}

\newcommand{\dsgd}{D-SGD\xspace}

\definecolor{tab10_blue}{rgb}{0.121, 0.466, 0.705}
\definecolor{tab10_orange}{rgb}{1.0,   0.498, 0.054}
\definecolor{tab10_green}{rgb}{0.172, 0.627, 0.172}
\definecolor{tab10_red}{rgb}{0.839, 0.152, 0.156}
\definecolor{tab10_purple}{rgb}{0.580, 0.403, 0.741}
\definecolor{tab10_brown}{rgb}{0.549, 0.337, 0.294}
\definecolor{tab10_pink}{rgb}{0.890, 0.466, 0.760}
\definecolor{tab10_gray}{rgb}{0.498, 0.498, 0.498}
\definecolor{tab10_olive}{rgb}{0.737, 0.741, 0.133}
\definecolor{tab10_cyan}{rgb}{0.090, 0.745, 0.811}

\input{lib/comments}

\newcommand*{\at}[1]{\ifthenelse{\equal{#1}{\star}}{^\star}{^{(#1)}}}
\newcommand*{\idx}[1]{_{#1}}
\newcommand*{\atidx}[2]{\at{#1}\idx{#2}}

%% file: lib/comments.tex
\def\commentType{1}

\ifnum\commentType=0
  \usepackage[disable]{todonotes} 

\fi
\ifnum\commentType=1

\fi
\ifnum\commentType=2
  \usepackage{pdfcomment}

\fi
\ifnum\commentType=3
  \usepackage{todonotes} 
  \usepackage{silence}
\fi

%% file: lib/math_macros.tex
\DeclarePairedDelimiterX{\abs}[1]{\lvert}{\rvert}{#1}
\DeclarePairedDelimiterX{\norm}[1]{\lVert}{\rVert}{#1}
\DeclarePairedDelimiterX{\cbr}[1]{\{}{\}}{#1} 
\DeclarePairedDelimiterX{\rbr}[1]{(}{)}{#1} 
\DeclarePairedDelimiterX{\sbr}[1]{[}{]}{#1} 

\DeclareMathOperator{\Var}{Var}

\providecommand{\R}{\mathbb{R}} 

\DeclareMathOperator{\expect}{\mathbb{E}}

\DeclareMathOperator{\sgn}{sign}
\makeatletter
\def\sign{\@ifnextchar*{\@sgnargscaled}{\@ifnextchar[{\sgnargscaleas}{\@ifnextchar{\bgroup}{\@sgnarg}{\sgn} }}}
\def\@sgnarg#1{\sgn\rbr{#1}}
\def\@sgnargscaled#1{\sgn\rbr*{#1}}
\def\@sgnargscaleas[#1]#2{\sgn\rbr[#1]{#2}}
\makeatother

\providecommand{\0}{\mathbf{0}}

\providecommand{\dd}{\mathbf{d}}

\let\ggg\gg
\renewcommand{\gg}{\mathbf{g}}

\providecommand{\xx}{\mathbf{x}}
\providecommand{\yy}{\mathbf{y}}
\providecommand{\zz}{\mathbf{z}}

\providecommand{\mI}{\mathbf{I}}

\providecommand{\mL}{\mathbf{L}}
\providecommand{\mM}{\mathbf{M}}

\providecommand{\mW}{\mathbf{W}}

\providecommand{\cL}{\mathcal{L}}

\providecommand{\cN}{\mathcal{N}}
\providecommand{\cO}{\mathcal{O}}

\providecommand{\one}{\mathds{1}}

\providecommand{\sqnorm}[1]{\norm{#1}^2}



%% file: 010_abstract.tex
\begin{abstract}
	In data-parallel optimization of machine learning models, workers collaborate to improve their estimates of the model: more accurate gradients allow them to use larger learning rates and optimize faster.
	In the decentralized setting, in which workers communicate over a sparse graph, current theory fails to capture important aspects of real-world behavior.
	First, the `spectral gap' of the communication graph is not predictive of its empirical performance in (deep) learning.
	Second, current theory does not explain that collaboration enables \emph{larger} learning rates than training alone.
	In fact, it prescribes \emph{smaller} learning rates, which further decrease as graphs become larger, failing to explain convergence dynamics in infinite graphs.
	This paper aims to paint an accurate picture of sparsely-connected distributed optimization.
	We quantify how the graph topology influences convergence in a quadratic toy problem and provide theoretical results for general smooth and (strongly) convex objectives.
	Our theory matches empirical observations in deep learning, and accurately describes the relative merits of different graph topologies. This paper is an extension of the conference paper by~\citet{vogels22beyondSpectral}. Code: \texttt{\href{https://github.com/epfml/topology-in-decentralized-learning}{github.com/epfml/topology-in-decentralized-learning}}.
\end{abstract}

%% file: 020_introduction.tex
\section{Introduction}
\label{sec:intro}

Distributed data-parallel optimization algorithms help us tackle the increasing complexity of machine learning models and of the data on which they are trained.
We can classify those training algorithms as either \emph{centralized} or \emph{decentralized}, and we often consider those settings to have different benefits over training `alone'.
In the \emph{centralized} setting, workers compute gradients on independent mini-batches of data, and they average those gradients between all workers.
The resulting lower variance in the updates enables larger learning rates and faster training.
In the \emph{decentralized} setting, workers average their models with only a sparse set of `neighbors' in a graph instead of all-to-all, and they may have private datasets sampled from different distributions.
As the benefit of decentralized learning, we usually focus only on the (indirect) access to other worker's datasets, and not of faster training.

\begin{figure}
    \includegraphics{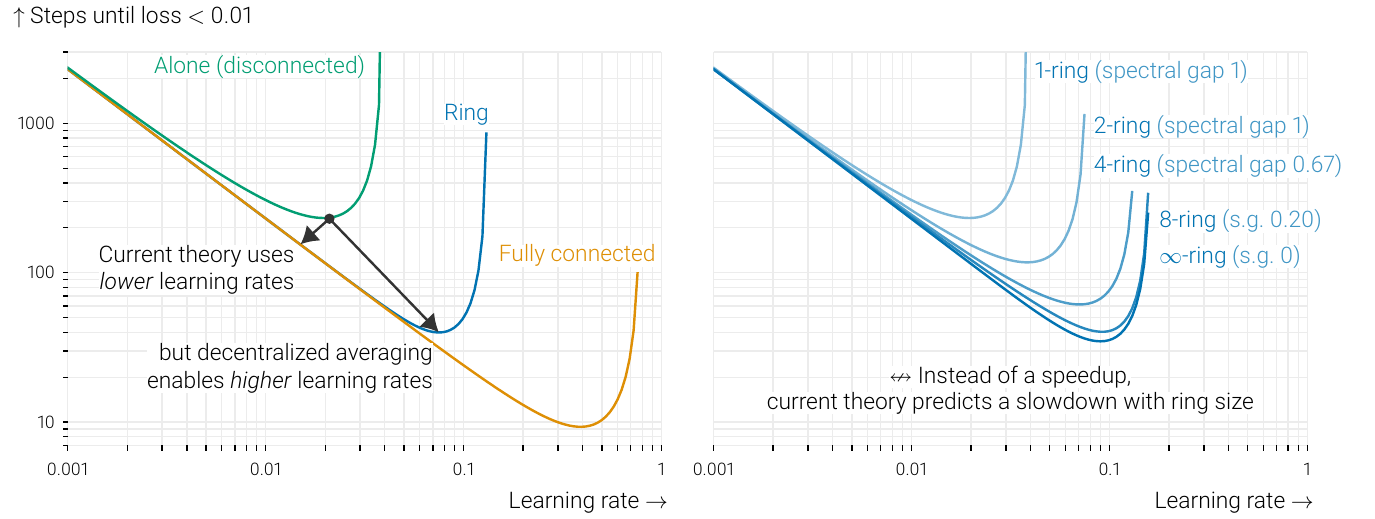}
    \centering
    \caption{
        \label{fig:intro_illustration}
        `Time to target' for \dsgd~\citep{lian2017can} with constant learning rates on an \iid isotropic quadratic dataset (\autoref{sec:toy-model}).
        The noise disappears at the optimum.
        Compared to optimizing alone, 32 workers in a ring (\emph{left}) are faster for any learning rate, but the largest improvement comes from being able to use a large learning rate.
        This benefit is not captured by current theory, which prescribes a smaller learning rate than training alone.
        On the \emph{right}, we see that rings of increasing size enable larger learning rates and faster optimization.
        Because a ring's spectral gap goes to zero with the size of the ring, this cannot be explained by current theory.
    }
\end{figure}

\emph{Homogeneous (i.i.d.)~setting.} While decentralized learning is typically studied with heterogeneous datasets across workers, sparse (decentralized) averaging between them is also useful when worker's data is identically distributed (i.i.d.)~\citep{lu2021optimal}.
As an example, sparse averaging is used in data centers to mitigate communication bottlenecks~\citep{assran2019sgp}.
When the data is \iid (or heterogeneity is mild), the goal of sparse averaging is to optimize faster, just like in centralized (all-to-all) graphs.
Yet, current decentralized learning theory poorly explains this speed-up.
Analyses typically show that, for \emph{small enough} learning rates, training with sparse averaging behaves the same as with all-to-all averaging~\citep{lian2017can,koloskova2020unified} and so it reduces the gradient variance by the number of workers compared to training alone with the \emph{same small learning rate}.
In practice, however, such small learning rates would never be used.
In fact, a reduction in variance should allow us to use a \emph{larger} learning rate than training alone, rather than imposing a \emph{smaller} one.
Contrary to current theory, we show that (sparse) averaging lowers variance throughout all phases of training (both initially and asymptotically), allowing to take higher learning rates, which directly speeds up convergence.
We characterize how much averaging with various communication graphs reduces the variance, and show that centralized performance (variance divided by the number of workers) is not always achieved when using optimal large learning rates.
The behavior we explain is illustrated in \autoref{fig:intro_illustration}.

\emph{Heterogeneous (non-i.i.d.)~setting.} In standard analyses,  heterogeneity affects convergence in a very worst-case manner.
Standard guarantees intuitively correspond to the pessimistic case in which the most distant workers have the most different functions.
These guarantees are typically loose in the settings where workers have different finite datasets sampled \iid from the same distribution, or if each worker has a lot of diversity in its close neighbors.
In this work, we characterize the impact of heterogeneity together with the communication graph, enabling non-trivial guarantees even for infinite graphs under non-adversarial heterogeneity patterns.

\emph{Spectral gap.} In both the homogeneous and heterogeneous settings, the graph topology appears in current convergence rates through the \emph{spectral gap} of its averaging (gossip) matrix.
The spectral gap poses a conservative lower bound on how much one averaging step brings all worker's models closer together.
The larger, the better.
If the spectral gap is small, a significantly smaller learning rate is required to make the algorithm behave close to SGD with all-to-all averaging with the same learning rate.
Unfortunately, we experimentally observe that, both in deep learning and in convex optimization, the spectral gap of the communication graph is \emph{not predictive} of its performance under tuned learning rates.

The problem with the spectral gap quantity is clearly illustrated in a simple example.
Let the communication graph be a ring of varying size.
As the size of the ring increases to infinity, its spectral gap goes to zero since it becomes harder and harder to achieve consensus between all the workers.
This leads to the optimization progress predicted by current theory to go to zero as well.
In some cases, when the worker's objectives are adversarially heterogeneous in a way that requires workers to obtain information from all others, this is indeed what happens.
In typical cases, however, this view is overly pessimistic.
In particular, this view does not match the empirical behavior with \iid data.
With \iid data, as the size of the ring increases, the convergence rate actually \emph{improves}~(\autoref{fig:intro_illustration}), until it saturates at a point that depends on the problem.

In this work, we aim to accurately describe the behavior of distributed learning algorithms with sparse averaging, both in theory and in practice.
We aim to do so both in the high learning rate regime, which was previously studied in the conference version of this paper~\citet{vogels22beyondSpectral}, as well as in the small learning rate regime, in which we characterize the interplay between topology and data heterogeneity, as well as stochastic noise. 
\begin{itemize}
\item We quantify the role of the graph in a quadratic toy problem designed to mimic the initial phase of deep learning (\autoref{sec:toy-model}), showing that averaging enables a larger learning rate.
\item From these insights, we derive a problem-independent notion of `effective number of neighbors' in a graph that is consistent with time-varying topologies and infinite graphs, and is predictive of a graph's empirical performance in both convex and deep learning.
\item We provide convergence proofs for (strongly) convex objectives that do not depend on the spectral gap of the graph (\autoref{sec:theory}), and consider finer spectral quantities instead.
Our rates disentangle the homogeneous and heterogeneous settings, and highlight that all problems behave as if they were homogeneous when the iterates are far from the optimum.
\end{itemize}
At its core, our analysis does not enforce global consensus, but only between workers that are close to each other in the graph.
Our theory shows that sparse averaging provably enables larger learning rates and thus speeds up optimization.
These insights prove to be relevant in deep learning, where we accurately describe the performance of a variety of topologies, while their spectral gap does not (\autoref{sec:experiments}).

%% file: 030_related_work.tex
\section{Related work}
\label{sec:related_work}

\paragraph{Decentralized SGD.}
This paper studies decentralized SGD.
\citet{koloskova2020unified} obtain the tightest bounds for this algorithm in the general setting where workers optimize heterogeneous objectives.
They show that gossip averaging reduces the asymptotic variance suffered by the algorithm at the cost of a degradation (depending on the spectral gap of the gossip matrix) of the initial linear convergence term.
This key term does not improve through collaboration and gives rise to a \emph{smaller learning rate} than training alone.
Besides, as discussed above, this implies that optimization is not possible in the limit of large graphs, even in the absence of heterogeneity: for instance, the spectral gap of an infinite ring is zero, which would lead to a learning rate of zero as well.

These rates suggest that decentralized averaging speeds up the last part of training (dominated by variance), at the cost of slowing down the initial (linear convergence) phase.
Beyond the work of~\citet{koloskova2020unified}, many papers focus on \emph{linear speedup} (in the variance phase) over optimizing alone, and prove similar results in a variety of settings~\citep{lian2017can,tang2018d,lian2018asynchronous}.
All these results rely on the following insight: while linear speedup is only achieved for small learning rates, SGD eventually requires such small learning rates anyway (because of, \eg, stochastic noise, or non-smoothness).
This observation leads these works to argue that ``topology does not matter''.
This is the case indeed, but only for very small learning rates, as shown in \autoref{fig:intro_illustration}.
Besides, while linear speedup might be achievable indeed for very small learning rates, some level of variance reduction should be obtained by averaging for \emph{any} learning rate.
In practice, averaging speeds up both the initial \emph{and} last part of training and in a possibly non-linear way.
This is what we show in this work, both in theory and in practice.

Another line of work studies decentralized SGD under statistical assumptions on the local data.
In particular, \citet{richards2020graph} show favorable properties for \dsgd with graph-dependent implicit regularization and attain optimal statistical rates.
Their suggested learning rate does depend on the spectral gap of the communication network, and it goes to zero when the spectral gap shrinks.
\citet{richards2019optimal} also show that larger (constant) learning rates can be used in decentralized GD, but their analysis focuses on decentralized kernel regression.
Their analysis relies on statistical concentration of local objectives rather, while the analysis in this paper relies on the notion of local neighborhoods.

\paragraph{Gossiping in infinite graphs.}
An important feature of our results is that they do not depend on the spectral gap, and so they apply independently of the size of the graph.
Instead, our results rely on new quantities that involve a combination of the graph topology and the heterogeneity pattern.
These may depend on the spectral gap in extreme cases, but are much better in general.
\citet{berthier20acceleratedgossip} study acceleration of gossip averaging in infinite graphs, and obtain the same conclusions as we do:
although spectral gap is useful for asymptotics (how long does information take to spread in the whole graph), it fails to accurately describe the transient regime of gossip averaging, \emph{i.e.}, how quickly information spreads over local neighborhoods in the first few gossip rounds.
This is especially limiting for optimization (compared to just averaging), as new local updates need to be averaged at every step.
The averaging for latest gradient updates always starts in the transient regime, implying that the transient regime of gossip averaging deeply affects the asymptotic regime of decentralized SGD.
In this work, we build on tools from~\citet{berthier20acceleratedgossip} to show how the effective number of neighbors, a key quantity we introduce, is related to the graph's spectral dimension.

\paragraph{The impact of the graph topology.}
\citet{lian2017can} argue that the topology of the graph does not matter.
This is only true for asymptotic rates in specific settings, as illustrated in \autoref{fig:intro_illustration}.
\citet{neglia20doestopomatter} investigate the impact of the graph on decentralized optimization, and contradict this claim.
Similarly to us, they show that the graph has an impact in the early phases of training.
Their analysis of the heterogeneous setting, their analysis depends on how gradient heterogeneity spans the eigenspace of the Laplacian.
Their assumptions, however, differ from ours, and they retain an unavoidable dependence on the spectral gap of the graph.
Our results are different in nature, and show the benefits of averaging and the impact of the graph through the choice of large learning rates, and a better dependence on the noise and the heterogeneity for a given learning rate.
\citet{even2021decentralized} also consider the impact of the graph on decentralized learning.
They focus on non-worst-case dependence on heterogeneous delays, and still obtain spectral-gap-like quantities but on a reweighted gossip matrix.

Another line of work studies the interaction of topology with particular patterns of data heterogeneity~\citep{lebars2022yes,dandi2022data}, and how to optimize graphs with this heterogeneity in mind.
Our analysis highlights the role of heterogeneity through a different quantity than these works, that we believe is tight.
Besides, both works either try to reduce this heterogeneity all along the trajectory, or optimize for both the spectral gap of the graph and the heterogeneity term.
Instead, we show that heterogeneity changes the fixed-point of the algorithm but not the global dynamics.

\paragraph{Time-varying topologies.}
Time-varying topologies are popular for decentralized deep learning in data centers due to their strong mixing~\citep{assran2019sgp,wang2019matcha}.
The benefit of varying the communication topology over time is not easily explained through standard theory, but requires dedicated analysis~\citep{ying2021exponential}.
While our proofs only cover static topologies, the quantities that appear in our analysis can be computed for time-varying schemes, too.
With these quantities, we can empirically study static and time-varying schemes in the same framework.

\paragraph{Conference version.}
This paper is an extension of~\citet{vogels22beyondSpectral}, which focused on the homogeneous setting where all workers share the same global optimum.
In this extension, we introduce a simpler analysis that strictly improves and generalizes the previous one, extending the results to the important heterogeneous setting.
In the conference version, it remained unclear if larger learning rates could only be achieved thanks to homogeneity.
We also connect the quantities we introduce to the spectral dimension of a graph, and use this connection to derive explicit formulas for the optimal learning rates based on the spectral dimension.
This allows us to accurately compare with previous bounds (for instance~\citet{koloskova2020unified}) and show that we improve on them in all settings.

%% file: 040_random_quadratics.tex
\section{Measuring collaboration in decentralized learning}

Both this paper's analysis of decentralized SGD for general convex objectives and its deep learning experiments revolve around a notion of `effective number of neighbors' that we would introduce in \autoref{sec:eff_nn}.
The aim of this section is to motivate the quantity based on a simple toy model for which we can exactly characterize the convergence (\autoref{sec:toy-model}).
We then connect this quantity to the typical graph metrics such as spectral gap and spectral dimensions in \autoref{sec:effnn_connections}.

\subsection{A toy problem: \dsgd on isotropic random quadratics}
\label{sec:toy-model}

The aim of this section is to provide intuition while avoiding the complexities of general analysis.
To keep this section light, we omit any derivations.
The appendix of \citep{vogels22beyondSpectral} contains a longer version of this section that includes derivations and proofs.

We consider $\nwrk$ workers that jointly optimize an isotropic quadratic $\expect_{\randdata\sim \cN^\dims(0,1)} \frac{1}{2}(\randdata^\top \param)^2 = \frac{1}{2}\norm{\param}^2$ with a unique global minimum $\param^\star = \0$.
The workers access the quadratic through stochastic gradients of the form $\grad(\param) = \randdata \randdata^\top \param$, with $\randdata \sim \cN^\dims(0, 1)$.
This corresponds to a linear model with infinite data, and where the model can fit the data perfectly, so that stochastic noise goes to zero close to the optimum.
We empirically find that this simple model is a meaningful proxy for the initial phase of (over-parameterized) deep learning (\autoref{sec:experiments}).
A benefit of this model is that we can compute exact rates for it.
These rates illustrate the behavior that we capture more generally in the theory of \autoref{sec:theory}.

The stochasticity in this toy problem can be quantified by the \emph{noise level} \begin{align} \noise = \sup_{\param \in \R^d} \frac{\expect_\randdata \norm{\grad(\param)}^2}{\norm{\param}^2} = \sup_{\param \in \R^d} \frac{\expect_\randdata \norm{\randdata \randdata^\top \param}^2}{\norm{\param}^2}, \end{align} which is equal to $\noise=\dims+2$, due to the random normal distribution of $\randdata$.

The workers run the \dsgd algorithm~\citep{lian2017can}.
Each worker $i$ has its own copy $\param\idx{i} \in \R^\dims$ of the model, and they alternate between local model updates $\param\idx{i} \gets \param\idx{i} - \lr\mspace{1mu}\grad(\param\idx{i})$ and averaging their models with others: $\param\idx{i} \gets \sum_{j=1}^\nwrk \gossipweight{ij} \param\idx{j}$.
The averaging weights $\gossipweight{ij}$ are summarized in the \emph{gossip matrix} $\gossip \in \R^{\nwrk\times \nwrk}$.
A non-zero weight $\gossipweight{ij}$ indicates that $i$ and $j$ are directly connected.
In the following, we assume that $\gossip$ is symmetric and doubly stochastic: $\sum_{j=1}^n \gossipweight{ij} = 1\;\forall i$.

On our objective, \dsgd either converges or diverges linearly.
Whenever it converges, \ie, when the learning rate is small enough, there is a convergence rate $\rate$ such that \vspace{-1mm} \begin{align*} \expect \norm{\param\atidx{t}{i}}^2 \leq (1 - \rate) \norm{\param\atidx{t-1}{i}}^2, \end{align*} with equality as $t \to \infty$.
When the workers train alone ($\gossip = \mI$), the convergence rate for a given learning rate $\lr$ reads: \vspace{-1mm} \begin{align} \label{eq:solo} \rate_\text{alone} = 1 - {\color{tab10_red}(1 - \lr)^2} - {\color{tab10_green}(\noise - 1) \lr^2}.
\end{align}\vspace{-1mm}%
The optimal learning rate $\lr^\star = \frac{1}{\noise}$ balances the optimization term $\color{tab10_red}(1-\lr)^2$ and the stochastic term $\color{tab10_green}(\noise - 1) \lr^2$.
In the centralized (fully connected) setting ($\gossipweight{ij} = \frac{1}{\nwrk}\; \forall i,j$), the rate is simple as well:\vspace{-2mm} \begin{align} \label{eq:centralized} \rate_\text{centralized} = 1 - {\color{tab10_red}(1 - \lr)^2} - \frac{\color{tab10_green}(\noise - 1) \lr^2}{\color{tab10_blue} \nwrk}.
\end{align}
Averaging between $\color{tab10_blue}\nwrk$ workers reduces the impact of the gradient noise, and the optimal learning rate grows to $\lr^\star = \frac{\color{tab10_blue}\nwrk}{{\color{tab10_blue}\nwrk}+\noise - 1}$.
We find that \dsgd with a general gossip matrix $\gossip$ interpolates those results.

\subsection{The effective number of neighbors}
\label{sec:eff_nn}
To quantify the reduction of the $\color{tab10_green}(\noise - 1) \lr^2$ term in general, we introduce the problem-independent notion of \emph{effective number of neighbors} $\nwrk_{\color{tab10_blue}\gossip}(\decay)$ of the gossip matrix $\color{tab10_blue}\gossip$ and \emph{decay parameter} $\decay$.

\begin{definition}[Effective number of neighbors]
    The effective number of neighbors $\nwrk_{\color{tab10_blue}\gossip}({\decay}) = \lim_{t\to\infty}  \frac{\sum_{i=1}^\nwrk\Var[\rwo\at{t}\idx{i}]}{\sum_{i=1}^\nwrk\Var[\rw\at{t}\idx{i}]}$ measures the ratio of the asymptotic variance of the processes
    \begin{align}
        \label{eqn:rw-noavg}
        \rwo\at{t+1} =\sqrt{{\decay}} \cdot \rwo\at{t} + \rwn\at{t},
        \quad \text{where } \rwo\at{t} \in \R^\nwrk \text{ and } \rwn\at{t} \sim \cN^\nwrk(0,1)
    \end{align}
    and
    \begin{align}
        \label{eqn:rw}
        \rw\at{t+1} = {\color{tab10_blue}\gossip}( \sqrt{{\decay}} \cdot \rw\at{t} + \rwn\at{t}),
        \quad \text{where } \rw\at{t} \in \R^\nwrk \text{ and } \rwn\at{t} \sim \cN^\nwrk(0,1).
    \end{align}
\end{definition}
We call $\rwo$ and $\rw$ \emph{random walks} because workers repeatedly add noise to their state, somewhat like SGD's parameter updates.
This should not be confused with a `random walk' over nodes in the graph.

Since averaging with ${\color{tab10_blue}\gossip}$ decreases the variance of the random walk by at most $\nwrk$, the effective number of neighbors is a number between $1$ and $\nwrk$.
The decay $\decay$ modulates the sensitivity to communication delays.
If $\decay=0$, workers only benefit from averaging with their direct neighbors.
As $\decay$ increases, multi-hop connections play an increasingly important role.
As $\decay$ approaches 1, delayed and undelayed noise contributions become equally weighted, and the reduction tends to $\nwrk$ for any connected topology.

\begin{proposition}
    \label{prop:eff_nn}
    For regular doubly-stochastic symmetric gossip matrices $\color{tab10_blue}\gossip$ with eigenvalues ${\color{tab10_blue}\weigval}_1, \ldots, {\color{tab10_blue}\weigval}_\nwrk$, $\nwrk_{\color{tab10_blue}\gossip}({\decay})$ has a closed-form expression\vspace{-2mm} \begin{align} \nwrk_{\color{tab10_blue}\gossip}({\decay}) = \frac{\frac{1}{1-{\decay}}}{\frac{1}{n}\sum_{i=1}^\nwrk \frac{{\color{tab10_blue}\weigval\idx{i}}^2}{1 - {\color{tab10_blue}\weigval}\idx{i}^2 {\decay}}}.
        \label{eq:var-reduction}
    \end{align}
\end{proposition}
This follows from unrolling the recursions for $\rwo$ and $\rw$, using the eigendecomposition of $\color{tab10_blue}\gossip$, and the limit $\lim t \to \infty \sum_{k=1}^{t} x^k = \frac{x}{1-x}$.

\begin{figure}
    \centering
    \includegraphics{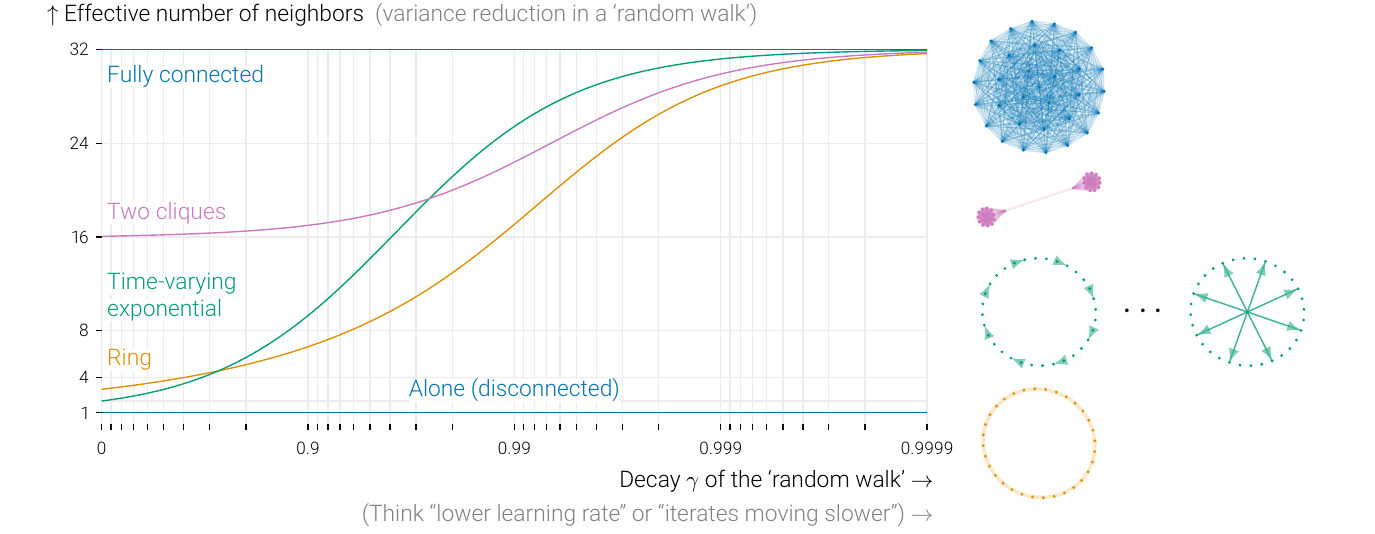}
    \vspace{-17pt}
    \caption{
        \label{fig:eff_nn}
        The effective number of neighbors for several topologies measured by their variance reduction in \eqref{eqn:rw}.
        The point $\decay$ on the $x$-axis that matters depends on the learning rate and the task.
        Which topology is `best' varies from problem to problem.
        For large decay rates $\decay$ (corresponding small learning rates), all connected topologies achieve variance reduction close to a fully connected graph.
        For small decay rates (large learning rates), workers only benefit from their direct neighbors (\eg 3 in a ring).
        These curves can be computed explicitly for constant topologies, and simulated efficiently for the time-varying exponential scheme~\citep{assran2019sgp}.
    }
\end{figure}

While this closed-form expression only covers a restricted set of gossip matrices, the notion of variance reduction in random walks, however, naturally extends to infinite topologies or time-varying averaging schemes.
\autoref{fig:eff_nn} illustrates $\nwrk_{\color{tab10_blue}\gossip}$ for various topologies.

In our exact characterization of the convergence of \dsgd on the isotropic quadratic toy problem, we find that the effective number of neighbors appears in place of the number of workers $\color{tab10_blue}\nwrk$ in the fully-connected rate of \autoref{eq:centralized}.
The rate $r$ is the unique solution to \begin{align} \label{eq:decentralized} \rate = 1 - {\color{tab10_red}(1 - \lr)^2} - \frac{\color{tab10_green}(\noise - 1) \lr^2}{\color{tab10_blue} \nwrk_\gossip\big(\frac{(1-\lr)^2}{1 - \rate}\big)}.
\end{align}
For fully-connected and disconnected $\color{tab10_blue}\gossip$, $\nwrk_{\color{tab10_blue}\gossip}({\decay})=\nwrk$ or 1 respectively, irrespective of $\decay$, and Equation~\ref{eq:decentralized} recovers Equations \ref{eq:solo} and \ref{eq:centralized}.
For other graphs, the effective number of workers depends on the learning rate.
Current theory only considers the case where ${\color{tab10_blue}\nwrk_\gossip} \approx \nwrk$, but the small learning rates this requires can make the term $\color{tab10_red} (1-\lr)^2$ too large, defeating the purpose of collaboration.

Beyond this toy problem, we find that the proposed notion of effective number of neighbors is also meaningful in the analysis of general objectives (\autoref{sec:theory}) and in deep learning (\autoref{sec:experiments}).

\subsection{Links between the effective number of neighbors and other graph quantities}
\label{sec:effnn_connections}

In general, the effective number of neighbors function $\neff$ cannot be summarized by a single scalar.
\autoref{fig:eff_nn} demonstrates that the behavior of this function varies from graph to graph.
We can, however, bound the effective number of neighbors by known graph quantities such as its spectral gap or spectral dimension.

We aim to create bounds for both finite and infinite graphs.
To allow for this, we introduce a generalization of Proposition~\ref{prop:eff_nn} as an integral over the spectral measure $\dd \sigma$ of the gossip matrix, instead of a sum over its eigenvalues: \begin{align} \nwrk_{\gossip}({\decay})^{-1} = (1-{\decay})\int_0^1 \frac{{\weigval}^2}{1 - \decay {\weigval}^2} \dd \sigma({\weigval}).
\end{align}
For finite graphs, $\dd \sigma$ is a sum of Dirac deltas of mass $\frac{1}{\nwrk}$ at each eigenvalue of matrix $\gossip$, recovering Equation~\eqref{eq:var-reduction}.

\subsubsection{Upper and lower bounds}
We can use the fact that there all eigenvalues $\weigval$ are $\leq 1$, leading to: \begin{equation} \nwrk_{\gossip}({\decay})^{-1} \leq (1-{\decay})\int_0^1 \frac{1}{1 - \decay} \dd \sigma({\weigval}) = 1, \end{equation} This lower bound to the `effective number of neighbors' corresponds to a disconnected graph.

On the other hand, for finite graphs, we can use the fact that $\sigma({\weigval})$ contains a series of $\nwrk$ Diracs.
The peak at $\weigval=1$, corresponding to the fully-averaged state, has value $\frac{1}{n}$, while the other peaks have values $\geq 0$.
Using this bound, we obtain \begin{equation} \nwrk_{\gossip}({\decay})^{-1} \geq \frac{1-{\decay}}{1 - \decay}\frac{1}{n} = \frac{1}{n}.
\end{equation}
This upper bound to the `effective number of neighbors' is tight for a fully-connected graph.

\subsubsection{Bounding by spectral gap}
If the graph has a spectral gap $\alpha$, this means that $\sigma(\weigval)$ contains a Dirac delta with mass $\frac{1}{n}$ at $\weigval=1$, corresponding to the fully-averaged state.
The rest of $\sigma(\weigval)$ has mass $\frac{n-1}{n}$ and is contained in the subdomain $\weigval \in [0, 1-\alpha]$.
In this setting, we obtain \begin{equation} \neff^{-1} \leq \frac{1}{n} + \frac{n- 1}{n} \frac{(1 - \decay)(1-\alpha)^2}{1 - \decay (1-\alpha)^2}.
\end{equation}
This lower bound to the `effective number of neighbors' is typically pessimistic, but it is tight for the finite gossip matrix $\gossip = (1-\alpha) \mI + \frac{\alpha}{n}\mathds{1}\mathds{1}^\top$.

\subsubsection{Bounding by spectral dimension}
Next, we will link the notion of `effective number of neighbors' to the spectral dimension $\spd$ of the graph~\citep[\eg Definition 1.9]{berthier2021analysis}, which controls the decay of eigenvalues near $1$.
This notion is usually linked with the spectral measure of the Laplacian of the graph.
However, to avoid introducing too many graph-related quantities, we define spectral dimension with respect to the gossip matrix $\gossip$.
Standard definitions using the Laplacian $\LW = \mI - \gossip$ are equivalent.
In the remainder of this paper, the `graph' will always refer to the communication graph implicitly induced by $\gossip$ of Laplacian $\LW$.

\begin{definition}[Spectral Dimension] \label{def:spec_measure}
    A gossip matrix has a spectral dimension at least $\spd$ if there exists $\CG > 0$ such that for all $\weigval \in [0,1]$, the density of its eigenvalues is bounded by \begin{equation} \label{eq:spectral_dimension} \sigma((\weigval, 1)) \leq \CG^{-1} (1 - \weigval)^\frac{\spd}{2}.
    \end{equation}
\end{definition}
The notation $\sigma((\weigval, 1))$ here refers to the integral $\int_\weigval^1 \sigma(l)\,\dd l$.
The spectral dimension of a graph has a natural geometric interpretation.
For instance, the line (or ring) are of spectral dimension $\spd = 1$, whereas 2-dimensional grids are of spectral dimension 2.
More generally, a $d$-dimensional torus is of spectral dimension $d$.
Besides, the spectral dimension describes macroscopic topological features and are robust to microscopic changes.
For instance, random geometric graphs are of spectral dimension 2.

Note that since finite graphs have a spectral gap, $\sigma((\lmax(\gossip),1)) = 0$ and so finite graphs verify~\eqref{eq:spectral_dimension} for any spectral dimension $\spd$.
However, the notion of spectral dimension is still relevant for finite graphs, since the constant $\CG$ blows up when $\spd$ is bigger than the actual spectral dimension of an infinite graph with similar topology.
Alternatively, it is sometimes helpful to explicitly take the spectral gap into account in~\eqref{eq:spectral_dimension}, as in~\citet[Section 6]{berthier20acceleratedgossip}.


We now proceed to bounding $\neff$ using the spectral dimension.
Since $ \weigval \mapsto \weigval^2(1 - \decay \weigval^2)^{-1}$ is a non-negative non-decreasing function on $[0, 1]$, we can use~\citet[Lemma C.1]{berthier20acceleratedgossip} to obtain that: \begin{equation} \label{eq:upper_bound_spectral_dim} \nwrk_{\gossip}({\decay})^{-1} \leq \frac{1}{\nwrk} + \CG^{-1} (1-{\decay}) \int_0^1 \frac{\weigval^2}{1 - \decay \weigval^2} (1 - \weigval)^{\frac{\spd}{2} - 1} \dd \weigval.
\end{equation}
The term $\frac{1}{n}$ comes from the fact that for finite graphs, the density $\dd \sigma$ includes a Dirac delta with mass $\frac{1}{n}$ at eigenvalue $1$.
This Dirac is not affected by spectral dimension, and is required for consistency, as it ensures that $\neff \leq \nwrk$ for any finite graph.
To evaluate the integral, we then distinguish three cases.

\paragraph{Case $\spd > 2$.}
Since $\decay \weigval < 1$, then $1 - \weigval \leq 1 - \decay \weigval^2$.
In particular we use integration by parts to get: \begin{align*} \nwrk_{\gossip}({\decay})^{-1} - \nwrk^{-1} & \leq \CG^{-1} (1-{\decay}) \int_0^1 \weigval^2 (1 - \decay \weigval^2)^{\frac{\spd}{2} - 2} \dd \weigval \\ &\leq - \frac{(1-{\decay})\CG^{-1}}{2 \decay (\spd/2 - 1)}  \int_0^1 - 2 \decay \weigval (\spd/2 - 1) (1 - \decay \weigval^2)^{\frac{\spd}{2} - 2} \dd \weigval\\  &= \frac{(1-{\decay})\CG^{-1}}{\decay (\spd - 2)} \left[1 - (1 - \decay )^{\frac{\spd}{2} - 1} \right].
\end{align*}
This leads to a scaling of: \begin{equation} \label{eq:spd_dg2} \nwrk_{\gossip}({\decay}) \geq \left(\frac{1}{\nwrk} + \frac{(1-{\decay})}{\decay (\spd - 2)\CG}\right)^{-1}.
\end{equation}
For large enough $\nwrk$, we obtain the same scaling of $(1 - \decay)^{-1}$ as in the previous section, thus indicating that for networks that are well-enough connected ($\spd > 2$), the spectral dimension only affects the constants, and not the scaling in $\decay$.

\paragraph{Case $\spd=2$.}
When $\spd = 2$, only the primitive of the integrand changes, leading to: \begin{equation} \label{eq:spd_d2} \nwrk_{\gossip}({\decay}) \geq \left( \frac{1}{\nwrk} - \frac{(1-{\decay})\ln(1 - \decay)} {2\decay \CG}\right)^{-1} \end{equation}        \paragraph{Case $\spd < 2$.
    }
In this case, we start by splitting the integral as: \begin{align*} (1 - \decay)\int_0^1 \frac{\weigval^2 (1 - \weigval)^{\frac{\spd}{2} - 1}}{(1 - \decay \weigval^2)} \dd \weigval &=  (1 - \decay)\int_0^{\decay} \frac{\weigval^2 (1 - \weigval)^{\frac{\spd}{2} - 1}}{(1 - \decay \weigval^2)} \dd \weigval  +  (1 - \decay)\int_{\decay}^1 \frac{\weigval^2 (1 - \weigval)^{\frac{\spd}{2} - 1}}{(1 - \decay \weigval^2)} \dd \weigval \end{align*} For the first term, note that $\decay \weigval \leq 1$, so $(1 - \decay \weigval^2)^{-1} \leq (1 - \weigval)^{-1}$, leading to: \begin{align*} (1 - \decay)\int_0^\decay \frac{\weigval^2 (1 - \weigval)^{\frac{\spd}{2} - 1}}{(1 - \decay \weigval^2)} \dd \weigval & \leq (1 - \decay)\int_0^{\decay} (1 - \weigval)^{\frac{\spd}{2} - 2} \dd \weigval \\   &= \frac{2(1 - \decay)}{2 - \spd}\left[(1 - \decay)^{\frac{\spd}{2} - 1} - 1\right] \leq \frac{2}{2 - \spd}(1 - \decay)^\frac{\spd}{2}.
\end{align*}
For the second term, note that $\weigval^2 \leq 1$, so $(1 - \decay \weigval^2)^{-1} \leq (1 - \decay)^{-1}$, leading to:       \begin{equation} (1 - \decay)\int_{\decay}^1 \frac{\weigval^2 (1 - \weigval)^{\frac{\spd}{2} - 1}}{(1 - \decay \weigval^2)} \dd \weigval \leq \int_{ \decay}^1  (1 - \weigval)^{\frac{\spd}{2} - 1} \dd \weigval = \frac{2}{\spd}(1 - \decay)^\frac{\spd}{2}.
\end{equation}
In the end, we obtain that $\nwrk_{\gossip}({\decay})^{-1} - \frac{1}{\nwrk} \leq \frac{2}{\CG}\left[\frac{1}{2 - \spd} + \frac{1}{\spd}\right] (1 - \decay)^{\frac{\spd}{2}}$, and so: \begin{equation} \label{eq:spd_dl2} \nwrk_{\gossip}({\decay}) \geq \left(\frac{1}{\nwrk} + \frac{4 (1 - \decay)^\frac{\spd}{2}}{\spd(2 - \spd) \CG} \right)^{-1}.
\end{equation}
In this case, scaling in $\decay$ is impacted by the spectral dimension.
Better-connected graphs benefit more from higher $\decay$.

%% file: 060_theory.tex
\section{Convergence analysis}
\label{sec:theory}

\subsection{Notations and Definitions}
In the previous section, we have derived exact rates for a specific function.
Now we present convergence rates for general (strongly) convex functions that are consistent with our observations in the previous section.
We obtain rates that depend on the level of noise, the hardness of the objective, and the topology of the graph.
More formally, we assume that we would like to solve the following problem: \begin{equation} \label{eq:dist_problem} \min_{\theta \in \R^d} \sum_{i=1}^\nwrk f_i(\theta) = \min_{\param \in \R^{\nwrk d}, \param_i = \param_j} \sum_{i=1}^\nwrk f_i(\param_i).
\end{equation}
In this case, $\param_i \in \R^d$ represents the local variable of node $i$, and $\param \in \R^{nd}$ the stacked variables of all nodes.
We will assume the following iterations for \dsgd: \begin{equation}\label{eq:def_dsgd} \text{(D-SGD):} \qquad \param\atidx{t+1}{i} = \sum_{j=1}^\nwrk \gossipweight{ij} \param\atidx{t}{j} - \lr \nabla f_{\xi\atidx{t}{i}}(\param\atidx{t}{i}) \end{equation} where $f_{\xi\atidx{t}{i}}$ represent sampled data points and the gossip weights $\gossipweight{ij}$ are elements of $\gossip$.
Denoting $\LW = \mI - \gossip$, we rewrite this expression in matrix form as: \begin{equation}\label{eq:def_dsgd_mat} \param\atidx{t+1}{} =  \param\atidx{t}{} - \left[\lr \nabla F_{\xi\atidx{t}{}}(\param\atidx{t}{}) + \LW \param\atidx{t}{}\right], \end{equation} where $(\nabla F_{\xi\atidx{t}{}}(\xt))_i = \nabla f_{\xi\atidx{t}{i}}(\param\atidx{t}{i})$.
We abuse notations in the sense that $\gossip \in \R^{\nwrk d \times \nwrk d}$ is now the Kronecker product of the standard $\nwrk \times \nwrk$ gossip matrix and the $d \times d$ identity matrix.

This definition is a slight departure from the conference version of this work~\citep{vogels22beyondSpectral}, which alternated randomly between gossip steps and gradient updates instead of in turns.
The analysis of the randomized setting is still possible, but with heterogeneous objectives $\param_{i} \neq \sum_{j=1}^\nwrk \gossipweight{ij} \param_{j}$, even for the fixed points of \dsgd~\eqref{eq:def_dsgd}, and randomizing the updates adds undesirable variance.
Similarly, it is also possible to analyze the popular variant $\param\atidx{t+1}{} = \gossip [\param\atidx{t}{} - \lr \nabla F_{\xi\atidx{t}{}}(\param\atidx{t}{})]$, which locally averages the stochastic gradients before they are applied.
Yet, the \dsgd algorithm in~\eqref{eq:def_dsgd} allows communications and computations to be performed in parallel, and leads to a simpler analysis.
We analyze this model under the following assumptions, where $D_f(x,y) = f(x) - f(y) - \nabla f(y)^\top(x-y)$ denotes the Bregman divergence of $f$ between points $x$ and $y$.

\begin{assumption}\label{assumption:stochastic}
	The stochastic gradients are such that: (\textsc{i}) the sampled data points $\xi\atidx{t}{i}$ and $\xi\atidx{\ell}{j}$ are independent across times $t, \ell$ and nodes $i \neq j$.
	(\textsc{ii}) stochastic gradients are locally unbiased: $\expect{[ f_{\xi\atidx{t}{i}}]} = f_i$ for all $t, i$
	(\textsc{iii}) the objectives $f_{\xi\atidx{t}{i}}$ are convex and $\noise_\xi$-smooth for all $t,i$, with $\esp{\noise_\xi D_{f_\xi}(x, y)} \leq \noise D_f(x, y)$ for all $x,y$.
	(\textsc{iv}) all local objectives $f_i$ are $\mu$-strongly-convex for $\mu\ge0$ and $\smoothf$-smooth.
\end{assumption}

\paragraph{Large learning rates.}
The smoothness constant $\noise$ of the stochastic functions $f_{\xi}$ defines the level of noise in the problem (the lower, the better) in the transient regime.
The ratio $\noise / \smoothf$ compares the difficulty of optimizing with stochastic gradients to the difficulty with the true global gradient before reaching the `variance region' in which the iterates of \dsgd with a constant learning rate lie almost surely as $t \to \infty$.
This ratio is thus especially important in interpolating settings when all $f_{\xi\atidx{t}{i}}$ have the same minimum, so that the `variance region' is reduced to the optimum $\xstar$.
Assuming better smoothness for the global average objective than for the local functions is key to showing that averaging between workers allows for larger learning rates.
Without communication, convergence to the `variance region' is ensured for learning rates $\lr \leq 1/\noise$.
If $\noise \approx \smoothf$, there is little noise and cooperation only helps to reduce the final variance, and to get closer to the \emph{global} minimum (instead of just your own).
Yet, in noisy regimes ($\noise \ggg \smoothf$), such as in \autoref{sec:toy-model} in which $\noise = \dims+2 \ggg 1 = \smoothf$, averaging enables larger learning rates up to $\min(1/\smoothf, \nwrk/\noise)$, greatly speeding up the initial training phase.
This is precisely what we will prove in Theorem~\ref{thm:convex_general}.

\emph{If} the workers always remain close ($\param\idx{i} \approx \frac{1}{n} (\param\idx{1} + \ldots + \param\idx{n})\;\forall i$, or equivalently $\frac{1}{n} \bm{1}\bm{1}^\top \param \approx \param$), \dsgd behaves the same as SGD on the average parameter $\frac{1}{n}\sum_{i=1}^\nwrk \param\idx{i}$, and the learning rate depends on $\max(\noise / \nwrk, \smoothf)$, showing a reduction of variance by $\nwrk$.
Maintaining ``$\frac{1}{n}\bm{1}\bm{1}^\top \param \approx \param$'', however, requires a small learning rate.
This is a common starting point for the analysis of \dsgd, in particular for the proofs in \citet{koloskova2020unified}.
On the other extreme, if we do not assume closeness between workers, ``$\mI \mspace{1mu} \param \approx \param$'' always holds.
In this case, there is no variance reduction, but no requirement for a small learning rate either.
In \autoref{sec:toy-model}, we found that, at the optimal learning rate, workers are \emph{not} close to all other workers, but they \emph{are} close to others that are not too far away in the graph.

We capture the concept of `local closeness' by defining a neighborhood matrix $\mmat \in \R^{n\times n}$.
It allows us to consider semi-local averaging beyond direct neighbors, but without fully averaging with the whole graph.
We ensure that ``$\mmat \param \approx \param$'', leading to an improvement in the smoothness somewhere between $\noise$ (achieved alone) and $\noise / \nwrk$ (achieved when global consensus is maintained).
Each neighborhood matrix $\mmat$ implies a requirement on the learning rate, as well as an improvement in smoothness.

While we can conduct our analysis with any $\mmat$, those matrices that strike a good balance between the learning rate requirement and improved smoothness are most interesting.
Based on \autoref{sec:toy-model}, we therefore focus on a specific construction of matrices: We choose $\mmat$ as the covariance of a decay-$\decay$ `random walk process' with the graph, as in \eqref{eqn:rw}, meaning that \begin{equation}\label{eq:M_definition} \mmat  = (1 - \decay)\sum_{k=1}^\infty \decay^{k - 1} \gossip^{2k} = (1-\decay)\gossip^2 (\mI- \decay \gossip^2)^{-1}.
\end{equation}
Varying $\decay$ induces a spectrum of averaging neighborhoods from $\mmat = \gossip^2$ ($\decay = 0$) to $\mmat = \frac{1}{n}\bm{1}\bm{1}^\top$ ($\decay = 1$).
$\decay$ also implies an effective number of neighbors $\nwrk_\gossip(\decay)$: the larger $\decay$, the larger $\nwrk_\gossip(\decay)$. We make the following assumption on the neighborhood matrix $\mmat$:
\begin{assumption}\label{ass:mmat}
	The neighborhood matrix $\mmat$ is of the form of \eqref{eq:M_definition}, and all the diagonal elements have the same value, \emph{i.e.,} $\mweight{ii} = \mweight{jj}$ for all $i,j$.
\end{assumption}
Assumption~\ref{ass:mmat} implies that $\mweight{ii}^{-1} = \neff$: the effective number of neighbors defined in~\eqref{eq:var-reduction} is equal to the inverse of the self-weights of $\mmat$. This comes from the fact that the trace of $\mmat$ is equal to the sum of its eigenvalues.
Otherwise, all results that require Assumption~\ref{ass:mmat} hold by replacing $\neff$ with $\min_i \mweight{ii}^{-1}$. Besides this interesting relationship with the effective number of neighbors $\neff$, we will be interested in another spectral property of $\mmat$, namely the constant $\lmwsmooth$ (which only depends on $\decay$ through $\mmat$, but we make this dependence explicit), which is such that:
\begin{equation}
	\LM \preccurlyeq \lmwsmooth^{-1}\LW \gossip
\end{equation}
This constant can be interpreted as the strong convexity of the semi-norm defined by $\LW \gossip$ relatively to the one defined by $\LM$. Due to the form of $\mmat$, we have $ 1 - \lmax(\gossip) \leq \lmwsmooth \leq 1$, and the lower bound is tight for $\gamma \rightarrow 1$. However, the specific form of $\mmat$ (involving neighborhoods as defined by $\gossip$) and the use of $\decay < 1$ ensure a much larger constant $\lmwsmooth$ in general.

\paragraph{Fixed points of D-(S)GD.}
In~\citet{vogels22beyondSpectral}, we consider a homogeneous setting, in which $\expect{f_{\xi\atidx{t}{i}}} = f$ for all $i$.
We now go beyond this analysis, and consider a setting in which local functions $f_i$ might be different.
In this case, constant-learning-rate Decentralized Gradient Descent (the deterministic version of \dsgd) does not converge to the minimizer of the average function  but to a different one.
Let us now consider this fixed point $\xstarmat$, which verifies: \begin{equation} \label{eq:def_xstarmat} \lr \nabla F(\xstarmat) + \LW \xstarmat = 0.
\end{equation}
Note that $\xstarmat$  crucially depends on the learning rate $\lr$ (which we emphasize in the notation) and that it is generally not at consensus ($\LW \xstarmat \neq 0$). In the presence of stochastic noise, \dsgd will oscillate in a neighborhood (proportional to the gradients' variance) of this fixed point $\xstarmat$, and so from now on we will refer to $\xstarmat$ as the fixed point of \dsgd.

In the remainder of this section, we show that the results from~\citet{vogels22beyondSpectral} still hold as long as we replace the global minimizer $\xstar$ (solution of Problem~\eqref{eq:dist_problem}) by this fixed point $\xstarmat$.
More specifically, we measure convergence by ensuring the decrease of the following Lyapunov function: \begin{equation} \label{eq:lyapunov} \cL_t = \norm{\param\at{t} - \xstarmat}^2_\mmat + \wcons \norm{\param\at{t} - \xstarmat}^2_\LM = (1 - \wcons)\norm{\param\at{t} - \xstarmat}^2_\mmat + \wcons \norm{\param\at{t} - \xstarmat}^2, \end{equation} for some parameter $\wcons \in [0,1]$, and where $\LM = \mI - \mmat$.
Then, we will show how these results imply convergence to a neighborhood of $\xstarmat$, and that this neighborhood shrinks with smaller learning rates $\lr$. More specifically, the section unrolls as follows:
\begin{enumerate}
	\item Theorem~\ref{thm:convex_general} first proves a general convergence result to $\xstarmat$, the fixed point of D-(S)GD.
	\item Theorem~\ref{thm:true_dist} then bounds the distance to the true optimum for general learning rates.
	\item Corollary~\ref{corr:final} finally gives a full convergence result with optimized learning rates.
	      Readers interested in quickly comparing our results with  state-of-the art ones can skip to this result.
\end{enumerate}

\subsection{General convergence result}
Theorem~\ref{thm:convex_general} provides convergence rates for any choice of the parameter $\decay$ that determines the neighborhood matrix $\mmat$, and for any Lyapunov parameter $\wcons$.
The best rates are obtained for specific~$\decay$ and $\wcons$ that balance the benefit of averaging with the constraint it imposes on closeness between neighbors.
We will discuss these choices more in depth in the next section.

\vspace{6pt plus2pt minus2pt}

\begin{mdframed}
	\begin{theorem} \label{thm:convex_general}
		If Assumptions~\ref{assumption:stochastic} and~\ref{ass:mmat} hold and if $\lr$ is such that
		\begin{equation} \label{eq:lr_conditions_thm} \lr \leq \min\left(\frac{\lmwsmooth \wcons}{\smoothf}, \frac{1}{4\left(\left[\mzero + \wcons\right]\noise + \smoothf\right)}\right), \end{equation}
		then the Lyapunov function defined in~\eqref{eq:lyapunov} verifies the following: \begin{align*} \cL\at{t+1} &\leq (1 - \lr \mu) \cL\at{t} + \lr^2\sigma_\mmat^2, \end{align*} where $\sigma_\mmat^2 = 2[(1 - \wcons)\mzero + \wcons] \esp{\norm{\nabla F_{\xi_t}(\xstarmat) - \nabla F(\xstarmat)}^2}$.
	\end{theorem}
\end{mdframed}

This theorem shows convergence (up to a variance region) to the fixed point $\xstarmat$ of \dsgd, regardless of the `true' minimizer $\xstar$.
Although converging to $\xstarmat$ might not be ideal depending on the use case (but do keep in mind that $\xstarmat \rightarrow \xstar$ as $\lr$ shrinks), this is what \dsgd does, and so we believe it is important to start by stating this clearly.
The homogeneous case did not have this problem since $\xstarmat = \xstar$ for all $\lr$ for $\lr$ that implied convergence.

Parameter $\wcons \in [0, 1]$ is free, and it is often convenient to choose it as $\wcons = \lr \smoothf / \lmwsmooth$ to get rid of the first condition on $\lr$.
However, we present the result with a free parameter $\wcons$ since, as we will see in the remainder of this section, setting $\wcons = \mzero$ allows for simple corollaries.

\begin{proof}
	We now detail the proof, which is both a simplification and generalization of Theorem~IV from~\citet{vogels22beyondSpectral}.

	\paragraph{1 - General decomposition}
	We first analyze the first term in the Lyapunov \eqref{eq:lyapunov}, and use the fixed-point conditions of \eqref{eq:def_xstarmat} to write: \begin{equation} \begin{split}\esp{\norm{\param\at{t+1} - \xstarmat}^2_\mmat} &= \norm{\param\at{t} - \xstarmat}^2_\mmat + \norm{\lr \nabla F_{\xi_t}(\xt) + \LW \xt}^2_\mmat\\ & - 2\lr \left[\nabla F(\xt) - \nabla F(\xstarmat)\right]^\top \mmat (\param\at{t} - \xstarmat) - 2 \norm{\xt - \xstarmat}^2_{\LW \mmat}.
		\end{split}
	\end{equation}
	The second term is the same with $\mmat$ in place of $\mI$.

	\paragraph{2 - Error terms}
	We start by bounding the error terms, and use the optimality conditions to obtain: \begin{align*} &\esp{\norm{\lr \nabla F_{\xi_t}(\param\at{t}) + \LW \xt}^2_{\mmat}} \\ &= \esp{\norm{\lr \left[\nabla F_{\xi_t}(\param\at{t}) - \nabla F(\xstarmat)\right] + \LW (\xt - \xstarmat)}^2_{\mmat}}\\ &= \esp{\norm{\lr \left(\nabla F_{\xi_t}(\param\at{t}) - \nabla F_{\xi_t}(\xstarmat)\right) + \left[\lr \left(\nabla F_{\xi_t}(\xstarmat) - \nabla F(\xstarmat)\right) + \LW (\xt - \xstarmat)\right]}^2_{\mmat}}\\ &\leq 2 \lr^2 \esp{\norm{\nabla F_{\xi_t}(\param\at{t}) - \nabla F_{\xi_t}(\xstarmat)}_\mmat^2} + 2\lr^2 \esp{\norm{\nabla F_{\xi_t}(\xstarmat) - \nabla F(\xstarmat)}^2_\mmat} + 2\norm{\xt - \xstarmat}^2_{\LW \mmat \LW}, \end{align*} where the last inequality comes from the bias-variance decomposition.
	The second term corresponds to variance, whereas the first and last one will be canceled by descent terms.

	\paragraph{Stochastic gradient noise.}
	To bound the first term, we crucially use that stochastic noises are \emph{independent} for two different nodes, so in particular: 
	\begin{align*} \esp{\norm{\nabla F_{\xi_t}(\param\at{t}) - \nabla F_{\xi_t}(\xstarmat)}^2_\mmat} &= \mzero \esp{\norm{\nabla F_{\xi_t}(\param\at{t}) - \nabla F_{\xi_t}(\xstarmat)}^2} \\
	& \qquad + \norm{\nabla F(\param\at{t}) - \nabla F( \xstarmat)}^2_{\mmat - \mzero \mI}\\ &  \leq 2 \mzero \esp{ \noise_{\xi_t} D_{F_{\xi_t}}(\xstarmat, \xt)} + \norm{\nabla F(\param\at{t}) - \nabla F(\xstarmat)}^2\\ &\leq 2\left[\mzero \noise + \smoothf\right] D_F(\xt, \xstarmat), \end{align*} where we used that $\mmat \preccurlyeq \mI$, the $\smoothf$-cocoercivity of $F$, and the noise assumption, \ie, $\esp{\noise_{\xi_t} D_{F_{\xi_t}}} \leq \noise D_F$.
	The effective number of neighbors $\neff$ kicks in since Assumption~\ref{ass:mmat} implies that the diagonal of $\mmat$ is equal to $\mzero \mI$. Using independence again, we obtain: \begin{equation} \esp{\norm{\nabla F_{\xi_t}(\xstarmat) - \nabla F(\xstarmat)}^2_\mmat} = \mzero \esp{\norm{\nabla F_{\xi_t}(\xstarmat) - \nabla F(\xstarmat)}^2} \end{equation} Performing the same computations for the $\esp{\norm{\nabla F_{\xi_t}(\param\at{t}) - \nabla F(\xstarmat)}^2}$ term and adding consensus error leads to: \begin{equation}\label{eq:main_error} \begin{split} \esp{\norm{\lr \nabla F_{\xi_t}(\param\at{t}) + \LW \xt}^2_{(1 - \wcons)\mmat + \wcons \mI}} &\leq 4\left[\left[(1 - \wcons)\mzero + \wcons\right]\noise + (1 - \wcons)\smoothf\right] D_F(\xt, \xstarmat)\\ &+  2\lr^2 ((1 - \wcons)\mzero + \wcons)\esp{\norm{\nabla F_{\xi_t}(\xstarmat) - \nabla F(\xstarmat)}^2}\\ &+ 2\norm{\xt - \xstarmat}^2_{\LW \left[\mmat + \wcons \LM\right] \LW} \end{split} \end{equation} Here, the first term will be controlled by the descent obtained through the gradient terms, and the second one through communication terms.

	\paragraph{3 - Descent terms}
	\paragraph{Gradient terms}

	We first analyze the effect of all gradient terms.
	In particular, we use that $(1 - \wcons) \mmat + \wcons \mI = \mI - (1 - \wcons) \LM$.
	Then, we use that \begin{align*} &\left[\nabla F(\xt) - \nabla F(\xstarmat)\right]^\top (\param\at{t} - \xstarmat) = D_F(\xt, \xstarmat) + D_F(\xstarmat, \xt), \end{align*} and: \begin{align*} 2\left[\nabla F(\xt) - \nabla F(\xstarmat)\right]^\top \LM (\param\at{t} - \xstarmat) & \leq 2\norm{\nabla F(\xt) - \nabla F(\xstarmat)} \norm{\LM (\param\at{t} - \xstarmat)} \\ &\leq \frac{1}{2\smoothf}\norm{\nabla F(\xt) - \nabla F(\xstarmat)}^2 + 2\smoothf \norm{\param\at{t} - \xstarmat}_{\LM^2} \\ &\leq D_F(\xt, \xstarmat) + 2\smoothf \norm{\param\at{t} - \xstarmat}_{\LM^2}.
	\end{align*}
	Overall, the gradient terms sum to: \begin{align} &- 2\left[\nabla F(\xt) - \nabla F(\xstarmat)\right]^\top (\param\at{t} - \xstarmat) + 2(1 - \wcons)\left[\nabla F(\xt) - \nabla F(\xstarmat)\right]^\top \LM (\param\at{t} - \xstarmat) \nonumber\\ &\leq - 2 D_F(\xstarmat, \xt) - (1 + \wcons) D_F(\xt, \xstarmat) + 2(1 - \wcons) \smoothf \norm{\param\at{t} - \xstarmat}_{\LM^2}\nonumber\\ &\leq - \mu \norm{\xt - \xstarmat}^2 - D_F(\xt, \xstarmat) + 2\smoothf \norm{\param\at{t} - \xstarmat}_{\LM^2}\nonumber\\ &\leq - (1 - \wcons)\mu \norm{\xt - \xstarmat}^2_\mmat - \wcons \norm{\xt - \xstarmat}^2 - D_F(\xt, \xstarmat) + 2\lmwsmooth^{-1}\smoothf \norm{\param\at{t} - \xstarmat}_{\LM \LW \gossip} \label{eq:main_descent_comp} , \end{align} where we used that $\LM \preccurlyeq \lmwsmooth^{-1}\LW \gossip$.

	\paragraph{Gossip terms.}
	We simply recall the gossip terms we use for descent here, which write: \begin{equation} \label{eq:main_descent_comm} - 2 \norm{\xt - \xstarmat}^2_{\LW \mmat} - 2 \wcons\norm{\xt - \xstarmat}^2_{\LW \LM}.
	\end{equation}

	\paragraph{4 - Putting everything together.}
	We now add all the descent and error terms together.
	More specifically, using Equations~\eqref{eq:main_error},~\eqref{eq:main_descent_comp} and \eqref{eq:main_descent_comm} we obtain: \begin{align*} \cL\at{t+1} & \leq (1 - \lr \mu)\cL\at{t} \\ &- 2 \norm{\xt - \xstarmat}^2_{\LW \mmat(\mI - \LW)} \\ &- 2\wcons \left[1 - \lr \smoothf / (\wcons\lmwsmooth)\right]\norm{\xt - \xstarmat}^2_{\LW \LM \gossip} \\ &- \lr\left(1 - 4\lr\left[\left[(1 - \wcons)\mzero + \wcons\right]\noise + (1 - \wcons)\smoothf\right] \right) D_F(\xt, \xstarmat)\\ &+ 2\lr^2 \left[(1 - \wcons)\mzero + \wcons\right] \esp{\norm{\nabla F_{\xi_t}(\xstarmat) - \nabla F(\xstarmat)}^2}.
	\end{align*}
	The conditions in the theorem are chosen so that the terms from lines 3 and 4 are positive (which is automatically true for line 2), and using that $1 - \wcons \leq 1$ (since $\wcons$ is small anyway).
\end{proof}

\subsection{Main corollaries}

\subsubsection{Large learning rate: speeding up convergence for large errors}

We now investigate Theorem~\ref{thm:convex_general} in the case in which both the noise $\sigma^2$ and the heterogeneity $\sqnorm{\nabla F(\xstar)}_{\LW^\dagger}$ are small (compared to $\cL\atidx{0}{}$), and so we would like to have the highest possible learning rate in order to ensure fast decrease of the objective (which is consistent with Figure~\ref{fig:intro_illustration}).
Using \eqref{eq:lr_conditions_thm}, we obtain a rate for each parameter $\decay$ that controls the local neighborhood size (remember that $\lmwsmooth$ depends on $\decay$).
The task that remains is to find the $\decay$ parameter that gives the best convergence guarantees (the largest learning rate).
As explained before, one should never reduce the learning rate in order to be close to others, because the goal of collaboration (in this regime in which we are not affected by variance and heterogeneity) is to \emph{increase} the learning rate.

\begin{figure}
	\includegraphics{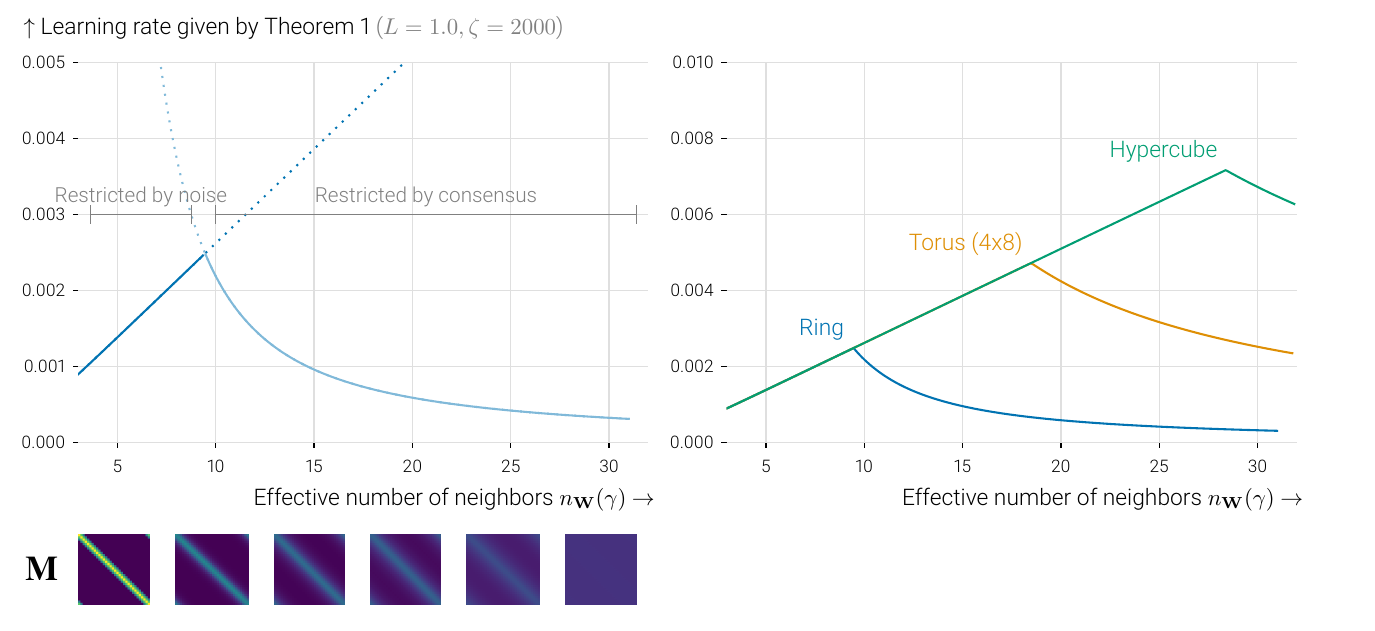}
	\centering
	\caption{ \label{fig:lr_gamma_theory}
		Maximum learning rates prescribed by \autoref{thm:convex_general}, varying the parameter $\decay$ that implies an effective neighborhood size ($x$-axis) and an averaging matrix $\mM$ (drawn as \emph{heatmaps}).
		On the \emph{left}, we show the details for a 32-worker ring topology, and on the \emph{right}, we compare it to more connected topologies.
		Increasing $\decay$ (and with it $\nwrk_\gossip(\decay)$) initially leads to larger learning rates thanks to noise reduction.
		At the optimum, the cost of consensus exceeds the benefit of further reduced noise.
	}
\end{figure}

We illustrate this in Figure~\autoref{fig:lr_gamma_theory}, that we obtain by choosing $\wcons = \mzero$, and evaluating the two terms of~\eqref{eq:lr_conditions_thm} for different values of $\decay$.
The expression for the linear part of the curve (before consensus dominates) is given in Corollary~\ref{cor:easy_lr}.



\begin{corollary}\label{cor:easy_lr}
	Consider that Assumptions~\ref{assumption:stochastic} and~\ref{ass:mmat} hold, then the largest (up to constants) learning rate is obtained as: \begin{equation} \lr = \left(8 \noise / \neff + 4 \smoothf\right)^{-1}, \text{ for $\decay$ such that } 4 \mzero \lmwsmooth (2\mzero\noise + \smoothf) \leq \smoothf \end{equation} \end{corollary} We see that the learning rate scales linearly with the number of effective neighbors in this case (which is equivalent to taking a mini-batch of size linear in $\neff$) until a certain number of neighbors is reached (condition on the right), or centralized performance is achieved ($\noise = \neff L$).
The condition on $\decay$ always has a solution since when $\decay\approx 0$, both $\lmwsmooth$ and $\mzero$ are close to $1$, and they both decrease when $\decay$ grows.
This corollary directly follows from taking $\wcons = \mzero$ in Theorem~\ref{thm:convex_general}.
Note that a slightly tighter choice could be obtained by setting $\wcons = \lr \lmwsmooth / L$.

\paragraph*{Investigating $\lmwsmooth$.}

We now evaluate $\lmwsmooth$ in order to obtain more precise bounds.
In particular, choosing $\mmat$ as in \eqref{eq:M_definition}, the eigenvalues of $\LM$ are equal to: \begin{equation} \lambda_i^{\LM} = \frac{1 - \lambda_i^2}{1 - \gamma \lambda_i^2}, \end{equation} where $\lambda_i$ are the eigenvalues of $\gossip$.
In particular, $\lmwsmooth \LM \preccurlyeq \gossip \LW$ translates into the fact that for all $i$ such that $\lambda_i \neq 1$ (automatically verified in this case), we want for all $i$: \begin{equation} \label{eq:beta_condition} \lmwsmooth \leq \frac{1 - \gamma \lambda_i^2}{1 - \lambda_i^2} (1 - \lambda_i) \lambda_i = \frac{\lambda_i(1 - \gamma \lambda_i^2)}{1 + \lambda_i}.
\end{equation}
We now make the simplifying assumption that $\lambda_{\min}(\gossip) \geq \frac{1}{2}$ (which we can always enforce by taking $\gossip^\prime = (\mI + \gossip) / 2$), but note that the theory holds regardless.
We motivate this simplifying assumption by the fact that the for arbitrarily small spectral gaps, the right side of \eqref{eq:beta_condition} will always be minimized for $\lmax(\gossip)$ assuming $\decay$ is large enough, so the actual value of $\lambda_{\min}(\gossip) < 1$ does not matter.
In particular, in this case, \emph{neglecting the effect of the spectral gap}, we can just take: \begin{equation} \label{eq:lmwsmooth} \lmwsmooth = \frac{1 - \decay \lmax(\gossip)}{4} \geq \frac{1 - \decay}{4}, \end{equation} Note that $\lmwsmooth$ allows for large $\decay$ when the spectral gap $1 - \lmax(\gossip)$ is large, but we allow non-trivial learning rates $\lr > 0$ even when $\lmax(\gossip) = 1$ (infinite graphs) as long as $\decay < 1$.

\paragraph*{Optimal choice of $\neff$.}
Leveraging the spectral dimension results from Section~\ref{sec:toy-model}, we obtain the following corollary: \begin{corollary} \label{cor:lr_spd} Under Assumption~\ref{assumption:stochastic} and~\ref{ass:mmat}, and assuming that $\lambda_{\min}(\gossip) \geq \frac{1}{2}$, that the communication graph has spectral dimension $\spd > 2$, and that $\noise \ggg \smoothf$, the highest possible learning rate is \begin{equation} \lr = \frac{1}{8}\left(\frac{\CG (\spd - 2)}{\noise^2 \smoothf}\right)^\frac{1}{3}, \text{ obtained for } \neff =  \left(\CG (\spd- 2) \frac{\noise}{\smoothf}\right)^{\frac{1}{3}} \end{equation} \end{corollary}  \noindent This result follows from Corollary~\ref{cor:easy_lr}, which, if $\noise \ggg \smoothf$, writes: \begin{equation} \frac{\smoothf}{\noise} \geq 8 \neff^{-2} \lmwsmooth = \neff^{-3} \CG(\spd - 2), \end{equation} where the right part is obtained by plugging in the expressions for $\lmwsmooth$ from~\eqref{eq:lmwsmooth} into $\neff^{-1} \leq \frac{2(1 - \decay)}{\CG (\spd - 2)}$ from~\eqref{eq:spd_dg2} (assuming $\decay \geq 1/2$).
Then, one can solve for $1 - \decay$.
Assumptions besides Assumption~\ref{assumption:stochastic} allow to give a simple result in this specific case, but similar expressions can easily be obtained for $\spd \leq 2$ and $\noise < \smoothf \neff$.

\subsubsection{Small learning rate: approaching the optimum arbitrarily closely}

Theorem~\ref{thm:convex_general} gives a convergence result to $\xstarmat$, the fixed point of \dsgd, and we have investigated in the previous section the behavior of \dsgd for large learning rates.
In Theorem~\ref{thm:true_dist}, we focus on small error levels, for which the \emph{variance} and \emph{heterogeneity} terms dominate, and we would like to take small learning rates $\lr$.
In this setting, we bound the distance between the current iterate and the \emph{true minimizer} $\xstar$ instead of $\xstarmat$.
We also provide a result that gets rid of all dependence on $\xstarmat$, and only explicitly depends on the learning rate $\lr$.

\begin{mdframed}
	\begin{theorem} \label{thm:true_dist}
		Under the same assumptions and conditions on the learning rate as Theorem~\ref{thm:convex_general} and Corollary~\ref{cor:lr_spd}, we have that: \begin{equation} {\color{tab10_red}\norm{\param\at{t} - \xstar}_\mmat} \leq 2(1 - \lr \mu)^t \cL\at{0} + {\color{tab10_blue}\frac{2\lr\sigma_\mmat^2}{\mu}} + {\color{tab10_orange}2\lr^2(1 + \kappa)\norm{\LW^\dagger \nabla F(\xstarmat)}^2} \end{equation} We can further remove $\xstarmat$ from the bound, and obtain: \begin{align*} {\color{tab10_red}\norm{\param\at{t} - \xstar}_\mmat} \leq 2(1 - \lr \mu)^t \cL\at{0} + {\color{tab10_blue}\frac{6\lr\sigma_{\mmat, \star}^2}{\mu}} + {\color{tab10_orange}6\lr^2 \kappa p^{-1} \hetero}, \end{align*} where ${\color{tab10_blue}\sigma_{\mmat, \star}^2} = (\mzero + \wcons) \esp{\sqnorm{\nabla F_\xi(\xstar) - \nabla F(\xstar)}}$ and ${\color{tab10_orange}p^{-1}} = \max_\lr \frac{\sqnorm{\LW^\dagger \nabla F(\xstarmat)}}{\sqnorm{\nabla F(\xstarmat)}_{\LW^\dagger}}$, so that $p \geq 1 - \lmax(\gossip)$, and ${\color{tab10_orange}\hetero} = \sqnorm{\nabla F(\xstar)}_{\LW^\dagger}$ \end{theorem} \end{mdframed}
		\noindent The norm $\color{tab10_red}\sqnorm{\xt - \xstar}_{\mmat}$ considers convergence of locally averaged neighborhoods, but $\sqnorm{\xt - \xstar}_{\mmat} \geq \| \overline{\param}\at{t} - \xstar\|^2$ since $\one$ is an eigenvector of $\mmat$ with eigenvalue $1$.
		We now briefly discuss the various terms in this corollary, and then prove it.

\paragraph{\color{tab10_orange}Heterogeneity term.}
The term due to heterogeneity only depends on the distance between the true optimum $\xstar$ and the fixed point $\xstarmat$, which we then transform into a condition on $\sqnorm{\nabla F(\xstar)}_{\LW^\dagger}$.
In particular, it is not influenced by the choice of $\mmat$ (and thus of $\decay$).

\paragraph{\color{tab10_orange}Constant $p$.}
We introduce constant $p$ to get rid of the explicit dependence on $\xstarmat$.
Indeed, $p^{-1}$ intuitively denotes how large $\LW^\dagger$ is in the direction of $\nabla F(\xstarmat)$.
For instance, if $\nabla F(\xstarmat)$ is an eigenvector of $\gossip$ associated with eigenvalue $\lambda$, then we have $p = 1 - \lambda$.
In the worst case, we have that $p = 1 - \lmax(\gossip)$, but $p$ can be much better in general, when the heterogeneity is spread evenly, instead of having very different functions on distant nodes.

\paragraph{\color{tab10_blue}Variance term.}
In this case, the largest variance reduction (of order $\nwrk$) is obtained by taking $\wcons$ and $\mzero$ as small as possible.
For learning rates that are too large to imply $\mzero \approx n^{-1}$, decreasing it decreases the variance term in two ways: \textsc{(i)} directly, through the $\lr$ term, \textsc{(ii)} indirectly, by allowing to take smaller values of $\mzero$.

For very large (infinite) graphs, we can take $\wcons = \mzero$, and in this case Theorem~\ref{thm:convex_general} gives that the smallest $\mzero$ is given by $\mzero \lmwsmooth = \lr \smoothf$.
Using spectral dimension results (for instance with $\spd > 2$), we obtain (similarly to Corollary~\ref{cor:lr_spd}) that we can take $\lmwsmooth = \mzero \CG (\spd - 2) / 8$, and so: \begin{equation} \mzero =  \sqrt{\frac{8\lr \smoothf}{\CG (\spd - 2)}}, \end{equation} so the residual variance term for this choice of $\mzero$ is of order: \begin{equation} \cO \left(\frac{\lr^{\frac{3}{2}}}{\mu} \sqrt{\frac{\smoothf}{\CG(\spd - 2)}}\esp{\sqnorm{\nabla F_\xi(\xstar) - \nabla F(\xstar)}} \right) \end{equation} In particular, we obtain \emph{super-linear} scaling when reducing the learning rate $\lr$ thanks to the added benefit of gaining more effective neighbors.
Note that again, the cases $\spd \leq 2$ can be treated in the same way.

\begin{proof}[Theorem~\ref{thm:true_dist}] We start by writing:
	\begin{equation}
		\sqnorm{\xt - \xstar}_\mmat \leq 2\sqnorm{\xt - \xstarmat}_\mmat + 2\sqnorm{\xstarmat - \xstar}_\mmat \leq 2 \cL\atidx{t}{} + 2 \sqnorm{\xstarmat - \xstar}.
	\end{equation}
	Theorem~\ref{thm:convex_general} ensures that $\cL\atidx{t}{}$ becomes small, and so we are left with bounding the distance between $\xstarmat$ and $\xstar$.

	\paragraph{1 - Distance to the global minimizer.}
	We define $\xstarmatbar = \one \one^\top \xstarmat / \nwrk$.
	Using the fact that both $\xstarmatbar$ and $\xstar$ are at consensus, and $\one^\top \nabla F(\xstarmat) = 0$ (immediate from~\eqref{eq:def_xstarmat}), we write: \begin{align} D_F(\xstar, \xstarmat) &= F(\xstar) - F(\xstarmat) - \nabla F(\xstarmat)^\top (\xstar - \xstarmat) \nonumber\\ &= F(\xstarmatbar) - F(\xstarmat) - \nabla F(\xstarmat)^\top (\xstarmatbar - \xstarmat) + F(\xstar) - F(\xstarmatbar)\nonumber\\ &\leq D_F(\xstarmatbar, \xstarmat), \label{eq:df_star_starmat} \end{align} where the last line comes from the fact that $\xstar$ is the minimizer of $F$ on the consensus space.
	Therefore: \begin{align*} \sqnorm{\xstarmat - \xstar} & = \sqnorm{\xstarmatbar - \xstar} + \sqnorm{\xstarmat - \xstarmatbar} \\ &\leq \frac{1}{\mu} D_F(\xstar, \xstarmat) + \sqnorm{\xstarmat - \xstarmatbar} \\ &\leq \frac{1}{\mu} D_F(\xstarmatbar, \xstarmat) + \sqnorm{\xstarmat - \xstarmatbar}\\ &\leq \left(1 + \frac{\smoothf}{\mu}\right) \sqnorm{\xstarmatbar - \xstarmat} = \lr^2 \left(1 + \frac{\smoothf}{\mu}\right)\sqnorm{ \LW^\dagger \nabla F(\xstarmat)}.
	\end{align*}
	Note that the result depends on the heterogeneity pattern of the gradients at the fixed point, and might be bounded (and even small) even when $\gossip$ has no spectral gap.
	However, this quantity is proportional to the squared inverse spectral gap in the worst case.

	\paragraph*{2 - Monotonicity in $\lr$.}
	We now prove that $\sqnorm{\nabla F(\xstarmat)}_{\LW^\dagger}$ decreases when $\lr$ increases, and so is maximal for $\lr = 0$, corresponding to $\xstarmat = \xstar$.
	More specifically:           \begin{align*} \frac{\dd\norm{\nabla F(\xstarmat)}^2_{\LW^\dagger }}{\dd \lr} &= \frac{\dd\left[ \lr^{-2}\norm{\xstarmat}^2_\LW\right]}{\dd \lr} = - \frac{2 \norm{\xstarmat}^2_\LW}{\lr^3} + 2 \lr^{-2} (\xstarmat)^\top \LW \frac{\dd\xstarmat}{\dd \lr} \end{align*} Differentiating the fixed-point conditions, we obtain that \begin{equation} \lr \nabla^2 F(\xstarmat) \dxstramat + \nabla F(\xstarmat) + \LW \dxstramat = 0, \end{equation} so that: \begin{equation} \dxstramat = - \left(\lr \nabla^2 F(\xstarmat) + \LW\right)^{-1} \nabla F(\xstarmat) = \lr^{-1}\left(\lr \nabla^2 F(\xstarmat) + \LW\right)^{-1} \LW \xstarmat.
	\end{equation}
	Plugging this into the previous expression and using that $\nabla^2 F(\xstarmat)$ is positive semi-definite, we obtain: \begin{align*} \frac{\dd\norm{\nabla F(\xstarmat)}^2_{\LW^\dagger}}{\dd \lr} & = - \frac{2}{\lr^3}(\xstarmat)^\top\left[\LW - \LW\left(\LW + \lr \nabla^2 F(\xstarmat)\right)^{-1} \LW \right] \xstarmat \\ &\leq - \frac{2}{\lr^3}(\xstarmat)^\top\left[\LW - \LW\LW^\dagger\LW \right] \xstarmat = 0.
	\end{align*}


	\paragraph*{3 - Getting rid of $\xstarmat$.}
	By definition of $p$, we can write: \begin{equation} \sqnorm{ \LW^\dagger \nabla F(\xstarmat)} \leq p^{-1} \sqnorm{\nabla F(\xstarmat)}_{\LW^\dagger} \leq p^{-1} \sqnorm{\nabla F(\xstar)}_{\LW^\dagger}.
	\end{equation}
	Note that we have to bound this constant $p$ in order to use the monotonicity in $\lr$ of $\sqnorm{\nabla F(\xstarmat)}_{\LW^\dagger}$ since this result does not hold for $\sqnorm{ \LW^\dagger \nabla F(\xstarmat)}$.
	For the variance, we write that: \begin{align*} \esp{\norm{\nabla F_{\xi_t}(\xstarmat) - \nabla F(\xstarmat)}^2} &\leq 3\esp{\sqnorm{\nabla F_{\xi_t}(\xstarmat) - \nabla F_{\xi_t}(\xstar)}} \\ &+ 3 \esp{\sqnorm{\nabla F_{\xi_t}(\xstar) - \nabla F(\xstar)}} + 3\sqnorm{\nabla F(\xstarmat) - \nabla F(\xstar)}\\ &\leq 3 \sigma^2_{\mmat, \star} + 3 \left(\noise + \smoothf\right) D_F(\xstar, \xstarmat). \end{align*} From here, we use Equation~\eqref{eq:df_star_starmat} and obtain that: \begin{equation} \esp{\norm{\nabla F_{\xi_t}(\xstarmat) - \nabla F(\xstarmat)}^2} \leq 3 \sigma^2_{\mmat, \star} + 3 \smoothf \left(\noise + \smoothf\right) \lr^2 \sqnorm{\LW^\dagger \nabla F(\xstarmat)}.
	\end{equation}
	To obtain the final result, we use that $\lr (\mzero + \wcons)(\noise + L) \leq 1/4$ thanks to the conditions on the learning rate.

\end{proof}

\subsubsection{Comparison with existing work.}
Expressed in the form of~\citet{koloskova2020unified}, we can summarize the previous corollaries into the following result by taking either $\lr$ as the largest possible constant (as indicated in Corollary~\ref{cor:lr_spd}) or $\lr = \tilde{O}(1/(\mu T))$.
Here, $\tilde{O}$ denotes inequality up to logarithmic factors, and recall that $\sqnorm{\xt - \xstar}_{\mmat} \geq \| \overline{\param}\at{t} - \xstar\|^2$.
We recall that $\smoothf$ is the smoothness of the global objective $f$, $\noise$ is the smoothness of the stochastic functions $f_\xi$, $\mu$ is the strong convexity parameter, $\spd$ is the spectral dimension of the gossip matrix $\gossip$ (and we assume $\spd > 2$) and $\CG$ is the associated constant.

\begin{mdframed}
	\begin{corollary}[Final result.] \label{corr:final}
		Under the same assumptions as Corollary~\ref{cor:lr_spd}, there exists a choice of learning rate (and, equivalently, of decay parameters $\decay_{\rm large}^*$ and $\decay_{\rm small}^*$) such that the expected squared distance to the global optimum after $T$ steps of \dsgd $\| \overline{\param}\at{t} - \xstar\|^2$ is of order: \begin{equation} \label{eq:corr_T_rate} \tilde{\cO}\left(\frac{\sigma^2}{\mu^2 T \nwrk_\gossip(\decay_{\rm small}^*)} + \frac{\smoothf \hetero}{\mu^3 p T^2} + \exp\left[- \nwrk_\gossip(\decay_{\rm large}^*) \frac{\mu}{\noise} T\right]\right), \end{equation} where $\hetero$ and $p$ are defined in Theorem~\ref{thm:true_dist}, and $\overline{\param}\at{t}$ is the average parameter.
		The optimal effective number of neighbors in respectively the small and large learning rate settings are: \begin{equation} \nwrk_\gossip(\decay_{\rm small}^*) = \min\left(\sqrt{\frac{\spd T}{\smoothf\CG}}, \nwrk\right) \text{ and } \nwrk_\gossip(\decay_{\rm large}^*) = \min\left(\left(\frac{\CG \spd \noise}{\smoothf}\right)^{\frac{1}{3}}, \nwrk\right).
		\end{equation}
	\end{corollary}
\end{mdframed}

This result can be contrasted with the result from~\citet{koloskova2020unified}, which writes: \begin{equation} \label{eq:corr_nastia} \tilde{\cO}\left(\frac{\sigma^2}{\mu^2 T}\left[ \frac{1}{n} + \frac{L}{\mu(1 - \lmax(\gossip))T}\right] + \frac{\smoothf \Delta^2}{\mu^3 (1 - \lmax(\gossip))^2 T^2} + \exp\left[- \frac{\mu}{(1 - \lmax(\gossip))\noise} T\right]\right), \end{equation}  We can now make the following observations.

\paragraph*{Scheduling the learning rate.} Here, the learning rate is either chosen as $\lr_{\rm large} = \nwrk_\gossip(\decay_{\rm large}^*) / \noise$, or as $\lr_{\rm small} = \tilde{O}((\mu T)^{-1})$. In practice, one would start with the large learning rate, and switching to $\lr_{\small}$ when training does not improve anymore (heterogeneity/variance terms dominate).

\paragraph*{Exponential decrease term.}
We first show a significant improvement in the exponential decrease term.
Indeed, $\nwrk_\gossip(\decay_{\rm large}^*) / (1 - \lmax(\gossip))$, the ratio between the largest learning rate permitted in our analysis versus existing ones, is always large since $\nwrk_\gossip(\decay_{\rm large}^*) \geq 1$ and $1 - \lmax(\gossip) \leq 1$.
Besides, the exponential decrease term is no longer affected by the spectral gap in our analysis, which only affects how big $\neff$ can be.
This improvement holds even when $\noise = \smoothf$ (in this case $\neff = 1$ is enough), and is due to the fact that \emph{heterogeneity only affects lower-order terms}, so that when cooperation brings nothing it doesn't hurt convergence either.

\paragraph*{Impact of heterogeneity.}
The improvement in the heterogeneous case does not depend on some $\decay$, and relies on bounding heterogeneity in a non-worst case fashion.
Indeed, $\noise_\gossip$ and $p$ capture the interplay between how heterogeneity is distributed among nodes, and the actual topology of the graph.
Note that this does not contradict the lower bound from~\citet{koloskova2020unified}, since $\hetero / p = \Delta^2 / (1 - \lmax(\gossip))^2$ in the worst case. In the worst case, the heterogeneity pattern of $\nabla F(\xstar)$ is aligned with the smallest eigenvalue of $\LW$, \emph{i.e.}, very distant nodes have very different objectives. The quantity $p$, however, gives more fine-grained bounds that depend on the actual heterogeneity pattern in general.

\paragraph*{Variance term.}
One key difference between the analyses is on the variance term that involves $\sigma^2$.
Both analyses depend on the variance of a single node, $\sigma^2 / (\mu T)$, which is then multiplied by a `variance reduction' term.
In both cases, this term is of the form $\neff^{-1} + \lr \smoothf \lmwsmooth^{-1}$.
However, the standard analysis implicitly use $\decay = 1$, and so $\neff = \nwrk$, and $\lmwsmooth = 1 - \lmax(\gossip)$.
Then, the form from~\eqref{eq:corr_nastia} follows from taking $\lr = \tilde{O}(1 / (\mu T))$.
Our analysis on the other hands relies on tuning $\decay$ such that $\neff^{-1} + \lr \smoothf \lmwsmooth^{-1}$ is the smallest possible, and is therefore strictly better than just considering $\decay = 1$.
Assuming a given spectral dimension $\spd > 2$ for the graph leads to~\eqref{eq:corr_T_rate}, but any assumption that precisely relates $\neff$ and $\decay$ would allow getting similar results.

While the $\tilde{O}(T^{-2})$ in the variance term of \citet{koloskova2020unified} seems better than our $\tilde{O}(T^{-3/2})$ term, this is misleading because constants are very important in this case.
Our rate is optimized by over $\decay$, which accounts for the fact that \emph{if} the $\tilde{O}(T^{-2})$ term dominates, then it is better to just consider a smaller neighborhood.
In that case, we would not benefit from $\nwrk^{-1}$ variance reduction anyway.
Our result optimally balances the two variance terms from~\eqref{eq:corr_nastia} instead.
Thanks to this balancing, we obtain that in graphs of spectral dimension $\spd > 2$, the variance decreases as $\tilde{O}(T^{-\frac{3}{2}})$ with a learning rate of $\tilde{O}(T^{-1})$ due to the combined effect of a smaller learning rate and adding more effective neighbors.
In finite graphs, this effect caps at $\neff = \nwrk$.

Finally, note that our analysis and the analysis of \citet{koloskova2020unified} allow for different generalizations of the standard framework: our analysis applies to arbitrarily large (infinite) graphs, while~\citet{koloskova2020unified} can handle time-varying graphs with weak (multi-round) connectivity assumptions.

%% file: 065_experiments.tex
\section{Empirical relevance in deep learning}
\label{sec:experiments}

While the theoretical results in this paper are for convex functions, the initial motivation for this work comes from observations in deep learning.
First, it is crucial in deep learning to use a large learning rate in the initial phase of training~\citep{li2019largelr}.
Contrary to what current theory prescribes, we do not use smaller learning rates in decentralized optimization than when training alone (even when data is heterogeneous.)
And second, we find that the spectral gap of a topology is not predictive of the performance of that topology in deep learning experiments.

In this section, we experiment with a variety of 32-worker topologies on Cifar-10~\citep{cifar10} with a VGG-11 model~\citep{simonyan2015vgg}.
Like other recent works~\citep{lin2021quasiglobal,vogels2021relay}, we opt for this older model, because it does not include BatchNorm~\citep{ioffe2015batchnorm} which forms an orthogonal challenge for decentralized SGD.
Please refer to Appendix E of \citep{vogels22beyondSpectral} for full details on the experimental setup.
Our set of topologies includes regular graphs like rings and toruses, but also irregular graphs such as a binary tree~\citep{vogels2021relay} and social network~\cite{davis1930socialwomen}, and a time-varying exponential scheme~\citep{assran2019sgp}.
We focus on the initial phase of training, 25k steps in our case, where both train and test loss converge close to linearly.
Using a large learning rate in this phase is found to be important for good generalization~\citep{li2019largelr}.

\newcommand{\kilo}{\hspace{1pt}k\xspace}

\begin{figure}
    \includegraphics{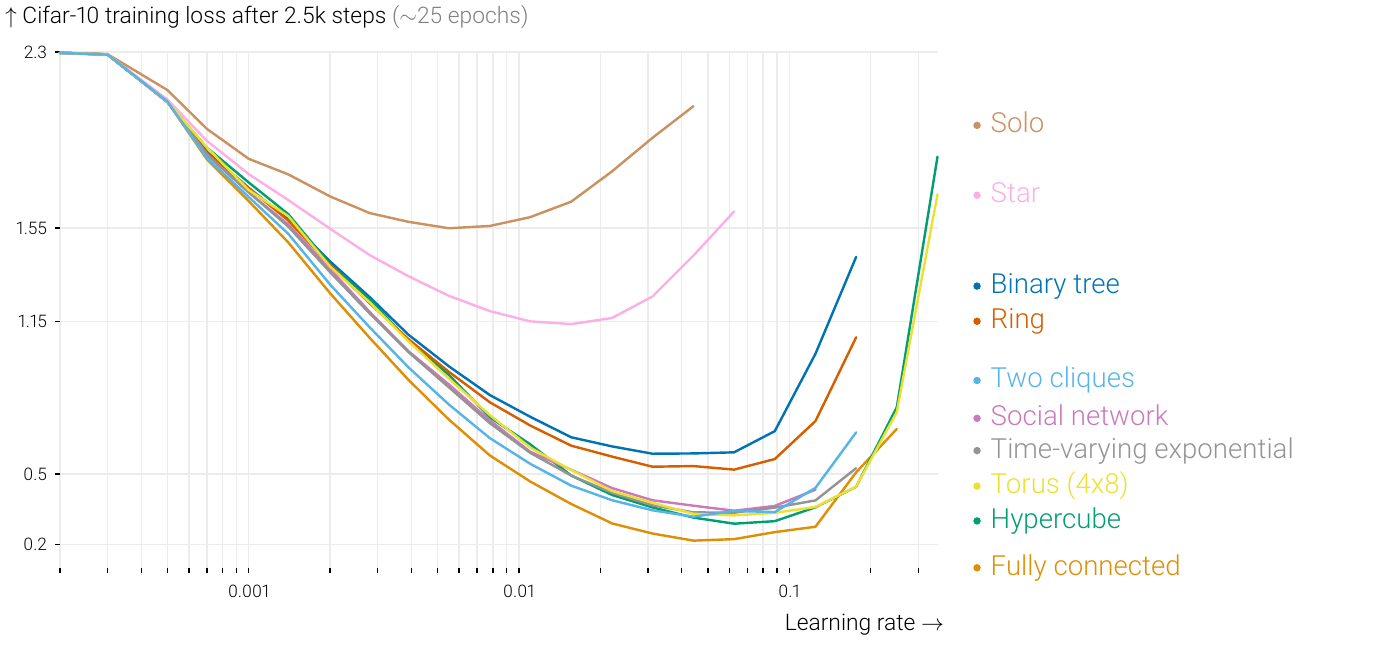}
    \centering
    \vspace{-18pt}
    \caption{
        \label{fig:results-cifar}
        Training loss reached after 2.5\kilo SGD steps with a variety of graph topologies.
        In all cases, averaging yields a small increase in speed for small learning rates, but a large gain over training alone comes from being able to increase the learning rate.
        While the star has a better spectral gap (0.031) than the ring (0.013), it performs worse, and does not allow large learning rates.
        For reference, similar curves for fully-connected graphs of varying sizes are in the appendix of~\citet{vogels22beyondSpectral}.
    }
\end{figure}

\autoref{fig:results-cifar} shows the loss reached after the first 2.5\kilo SGD steps for all topologies and for a dense grid of learning rates.
The curves have the same global structure as those for isotropic quadratics \autoref{fig:intro_illustration}: (sparse) averaging yields a small increase in speed for small learning rates, but a large gain over training alone comes from being able to increase the learning rate.
The best schemes support almost the same learning rate as 32 fully-connected workers, and get close in performance.

\begin{figure}
    \includegraphics[width=0.99\linewidth]{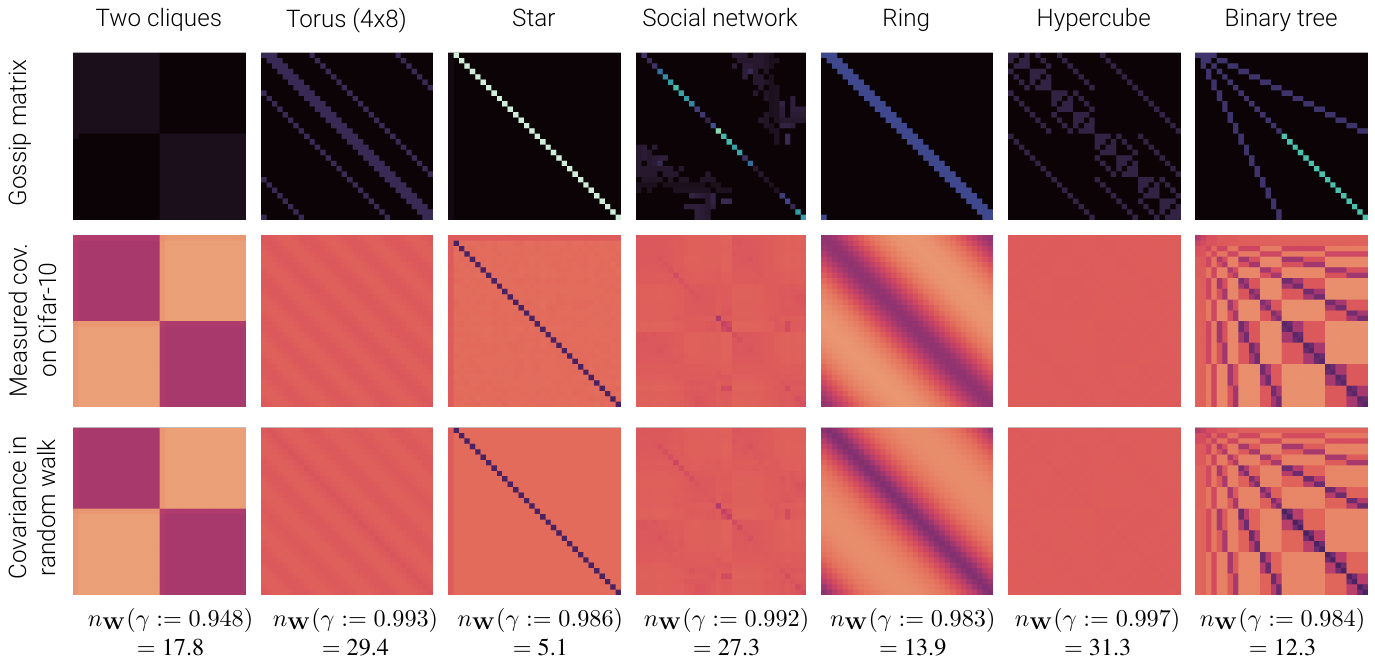}
    \vspace{-10pt}
    \centering
    \caption{
        \label{fig:cifar-covariance}
        Measured covariance in Cifar-10 (second row) between workers using various graphs (top row).
        After 10 epochs, we store a checkpoint of the model and train repeatedly for 100 SGD steps, yielding 100 models for 32 workers.
        We show normalized covariance matrices between the workers.
        These are very well approximated by the covariance in the random walk process of \autoref{sec:toy-model} (third row).
        We print the fitted decay parameters and corresponding `effective number of neighbors'.
        \vspace{-10pt}
    }
\end{figure}


We also find that the random walks introduced in \autoref{sec:toy-model} are a good model for variance between workers in deep learning.
\autoref{fig:cifar-covariance} shows the empirical covariance between the workers after 100 SGD steps.
Just like for isotropic quadratics, the covariance is accurately modeled by the covariance in the random walk process for a certain decay rate $\decay$.

\begin{figure}
    \includegraphics{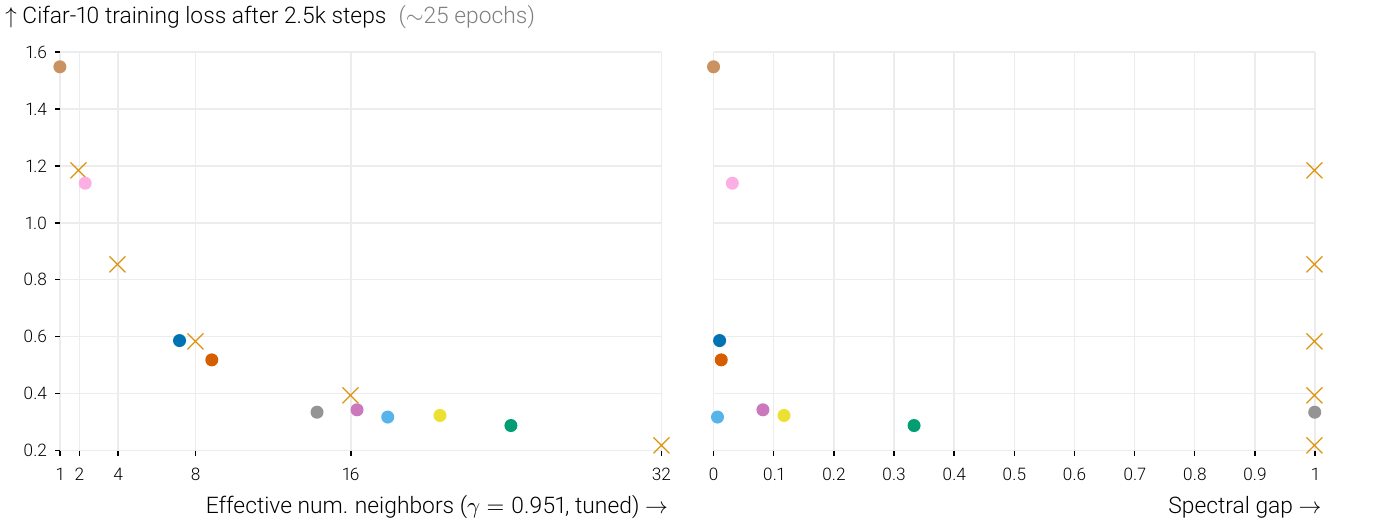}
    \centering
    \caption{
        Cifar-10 training loss after 2.5\kilo steps
        for all studied topologies with their optimal learning rates.
        Colors match \autoref{fig:results-cifar}, and {\color{tab10_orange}$\times$} indicates fully-connected graphs with varying number of workers.
        After fitting a decay parameter $\decay=0.951$ that captures problem specifics, the effective number of neighbors (left) as measured by variance reduction in a random walk (like in \autoref{sec:toy-model}) explains the relative performance of these graphs much better than the spectral gap of these topologies (right).
        \label{fig:correlation_plots}
    }
\end{figure}

Finally, we observe that the effective number of neighbors computed by the variance reduction in a random walk (\autoref{sec:toy-model}) accurately describes the relative performance under tuned learning rates of graph topologies on our task, including for irregular and time-varying topologies.
This is in contrast to the topology's spectral gaps, which we find to be not predictive.
We fit a decay rate $\decay=0.951$ that seems to capture the specifics of our problem, and show the correlation in \autoref{fig:correlation_plots}.

Appendix F of \citep{vogels22beyondSpectral} replicates the same experiments in a different setting.
There, we use larger graphs (of 64 workers), a different model and data set (an MLP on Fashion MNIST~\cite{xiao2017fashion}), and no momentum or weight decay.
The results in this setting are qualitatively comparable to the ones presented above.

%% file: 070_conclusion.tex
\section{Conclusion}

We have shown that the sparse averaging in decentralized learning allows larger learning rates to be used, and that it speeds up training.
With the optimal large learning rate, the workers' models are not guaranteed to remain close to their global average.
Enforcing global consensus is often unnecessary and the small learning rates it requires can be counter-productive.
Indeed, models \emph{do} remain close to some local average in a weighted neighborhood around them even with high learning rates.
The workers benefit from a number of `effective neighbors', potentially smaller than the whole graph, that allow them to use larger learning rates while retaining sufficient consensus within the `local neighborhood'. 

Similar insights apply when nodes have heterogeneous local functions: there is no need to enforce global averaging over the whole network when heterogeneity is small across local neighborhoods. Besides, there is no need to compensate for heterogeneity in the early phases of training, when models are all far from the global optimum.


Based on our insights, we encourage practitioners of sparse distributed learning algorithms to look beyond the spectral gap of graph topologies, and to investigate the actual `effective number of neighbors' that is used.
We also hope that our insights motivate theoreticians to be mindful of assumptions that artificially limit the learning rate, even though they are tight in worst cases. Indeed, the spectral gap is omnipresent in the decentralized litterature, which sometimes hides some subtle phenomena such as the superlinear decrease of the variance in the learning rate, that we highlight.  

We show experimentally that our conclusions hold in deep learning, but extending our theory to the non-convex setting is an important open direction that could reveal interesting new phenomena. Another interesting direction would be to better understand (beyond the worst-case) the effective number of neighbors for irregular graphs.

%% file: 080_acknowledgements.tex
\acks{This project was supported by SNSF grant 200020\_200342.

	We thank Lie He for valuable conversations and for identifying the discrepancy between a topology's spectral gap and its empirical performance.

	We also thank Rapha\"el Berthier for helpful discussions that allowed us to clarify the links between effective number of neighbors and spectral dimension.

	We also thank Aditya Vardhan Varre, Yatin Dandi and Mathieu Even for their feedback on the manuscript.}